\documentclass[10pt,journal,compsoc]{IEEEtran}
\ifCLASSOPTIONcompsoc
  \usepackage[nocompress]{cite}
\else
  \usepackage{cite}
\fi
\ifCLASSINFOpdf
\else
\fi
\usepackage{multirow}
\usepackage{subfigure}
\usepackage{algorithm}
\usepackage[noend]{algorithmic}
\usepackage{xspace}
\usepackage{dsfont}
\usepackage{epsfig}
\usepackage{epstopdf}
\usepackage{booktabs}
\usepackage{indentfirst}
\usepackage{amsmath}
\usepackage{amsthm}
\usepackage{bm}
\usepackage{listings}
\usepackage{tabu}
\usepackage{longtable}
\usepackage{threeparttable}
\usepackage{marvosym}
\usepackage{color}
\usepackage{tabularx}
\usepackage{hyperref}
\usepackage{xspace}
\usepackage{multirow}
\usepackage{subfigure}
\usepackage{algorithm}
\usepackage{xspace}
\usepackage{dsfont}
\usepackage{epsfig}
\usepackage{epstopdf}
\usepackage{booktabs}
\usepackage{indentfirst}
\usepackage{amsmath}
\usepackage{amsthm}
\usepackage{amsfonts}
\usepackage{bm}
\usepackage{listings}
\usepackage{tabu}
\usepackage{longtable}
\usepackage{threeparttable}
\usepackage{marvosym}
\usepackage{tabularx}
\usepackage{color}
\usepackage{enumerate}   
\usepackage{enumitem}
\usepackage{tcolorbox}

\newcommand{\name}{LibCity\xspace}
\newcommand{\modelCnt}{65\xspace}
\newcommand{\rawDataCnt}{55\xspace}
\newcommand{\paperCnt}{351\xspace}
\newcommand{\taskcnt}{9\xspace}

\newcommand{\ie}{\emph{i.e.,}\xspace}

\newcommand{\eg}{\emph{e.g.,}\xspace}

\newcommand{\ignore}[1]{}

\begin{document}
% \title{\name: A Unified and Extensible Library for Urban Spatial-Temporal Prediction: Design, Implementation, and Evaluation}
% \title{Urban Spatial-Temporal Data Management, Analysis, and Prediction with a Comprehensive Library \name}
\title{\name: A Unified Library Towards Efficient and Comprehensive Urban Spatial-Temporal Prediction}

\author{Jiawei Jiang,~
        Chengkai Han,~
        Wenjun Jiang,~
        Wayne Xin Zhao,~\IEEEmembership{Member,~IEEE},
        Jingyuan Wang,~\IEEEmembership{Member,~IEEE}% <-this % stops a space
\IEEEcompsocitemizethanks{\IEEEcompsocthanksitem Jingyuan Wang is with the School of Computer Science and Engineering, Beihang University, Beijing 100191, China, and also with School of Economics and Management, Beihang University, Beijing 100191, China. \protect\\
E-mail: jywang@buaa.edu.cn
\IEEEcompsocthanksitem Jiawei Jiang, Wenjun Jiang, and Chengkai Han are with the School of Computer Science and Engineering, Beihang University, Beijing 100191, China.
E-mail: \{jwjiang, jwjun, ckhan\}@buaa.edu.cn
\IEEEcompsocthanksitem Wayne Xin Zhao is with the Gaoling School of Artificial Intelligence, Renmin University of China, Beijing 100872, China
}% <-this % stops a space
\thanks{Corresponding Author: Jingyuan Wang.}}

\markboth{IEEE TRANSACTIONS ON KNOWLEDGE AND DATA ENGINEERING, ~Vol.~XX, No.~XX, May~2023}%
{Wang \MakeLowercase{\textit{et al.}}: \name: A Unified Library Towards Efficient and Comprehensive Urban Spatial-Temporal Prediction}



\IEEEtitleabstractindextext{%
\begin{abstract}
As deep learning technology advances and more urban spatial-temporal data accumulates, an increasing number of deep learning models are being proposed to solve urban spatial-temporal prediction problems. However, there are limitations in the existing field, including open-source data being in various formats and difficult to use, few papers making their code and data openly available, and open-source models often using different frameworks and platforms, making comparisons challenging. A standardized framework is urgently needed to implement and evaluate these methods. To address these issues, we propose \name, an open-source library that offers researchers a credible experimental tool and a convenient development framework. In this library, we have reproduced \modelCnt spatial-temporal prediction models and collected \rawDataCnt spatial-temporal datasets, allowing researchers to conduct comprehensive experiments conveniently. By enabling fair model comparisons, designing a unified data storage format, and simplifying the process of developing new models, \name is poised to make significant contributions to the spatial-temporal prediction field. % Using \name, we conducted a series of experiments to validate the effectiveness of different models and components, and we summarized promising future technology developments and research directions for spatial-temporal prediction. provide a comprehensive review of urban spatial-temporal prediction and propose a unified storage format for spatial-temporal data called \textit{atomic files}. We also 
\end{abstract}

\begin{IEEEkeywords}
Spatial-Temporal Prediction, Open-source Library
\end{IEEEkeywords}
}

\maketitle

\ifCLASSOPTIONcompsoc
\IEEEraisesectionheading{\section{Introduction}\label{intro}}
\else
\section{Introduction}
\label{intro}
\fi

\IEEEPARstart{I}{n} recent years, with the advancement of sensor technology in urban areas, a large amount of data can be collected, providing new perspectives for using artificial intelligence technologies to solve urban prediction problems~\cite{yin2021deep}. Solving spatial-temporal prediction problems is crucial in urban computing, facilitating the management and decision-making processes of smart cities and improving residents' living standards. The urban spatial-temporal prediction has numerous applications, including congestion control~\cite{control}, route planning~\cite{routeplanning}, vehicle dispatching~\cite{dispatching}, and POI recommendation~\cite{POIrec}.

Numerous urban spatial-temporal prediction techniques have been proposed in the literature. Unfortunately, we found that less than 30\% of the papers published in 11 leading conferences and journals have made their code and data open source, which creates reproducibility challenges in the field~\cite{libcity}. Additionally, these models are often implemented under different platforms or frameworks, which makes it challenging to reproduce the results in a unified manner for researchers. In particular, the accuracy of prediction models on a specific dataset is sensitive to the choice of hyperparameters. Without a public and unified standard dataset for benchmarking model performance, it is increasingly difficult to measure the effectiveness of new spatial-temporal prediction methods and fairly compare the performance of different models~\cite{yin2021deep, trafficsurvey1}. 

In contrast, domains like Computer Vision, Natural Language Processing, and Recommendation Systems have standardized datasets such as IMAGENET~\cite{IMAGENET} and algorithm libraries like MMDetection~\cite{mmdetection} and RecBole~\cite{recbole}. Unfortunately, urban spatial-temporal prediction lacks these resources. Therefore, we urgently need to develop a standardized library that considers all aspects of urban spatial-temporal prediction. To address these challenges, we introduce \name~\footnote{\url{https://github.com/LibCity}}, an open-source library that supports standardized measurement of models. By providing a standard library for urban spatial-temporal prediction, we aim to enhance reproducibility, comparability, and the advancement of this field.

The main features of \name can be summarized in four aspects: 

\begin{itemize}
    \item \textbf{Unified and Modular Framework Design}: \name adopts a comprehensive and standardized approach to implementing, deploying, and evaluating spatial-temporal prediction models. The library is built entirely on PyTorch~\cite{pytorch} and comprises five modules: Configuration, Data, Model, Evaluation, and Execution. Each module has a well-defined scope and collaborates seamlessly with others to deliver the complete functionality of the library. We design basic spatial-temporal data storage, unified model instantiation interfaces, and standardized evaluation procedure within these modules. Users can effortlessly train and evaluate existing models with simple configurations. On the other hand, developers can concentrate solely on the interfaces that are relevant to their models without worrying about the implementation details of other modules.
    \item \textbf{General and Extensible Data Storage Format}: Open-source spatial-temporal datasets are available in various storage formats. To provide a user-friendly interface and ensure the library's uniformity, \name has developed a general and extensible data storage format, namely \textit{atomic files}, for urban spatial-temporal data. These atomic files consist of five categories, which represent the minimum information units in spatial-temporal data, and include Geographical Unit Data, User Unit Data, Unit Relation Data, Spatial-temporal Dynamic Data, and External Data. The atomic files are a generic and extensible structure that enables the representation of spatial-temporal data consistently. Using atomic files, \name has developed data processing functions and \textit{Batch} extraction tools to create a unified data processing process, minimizing the effort required to develop new models. \textit{Batch} is the standardized input format for models in \name.
    \item \textbf{Comprehensive Benchmark Tasks, Datasets and Models}: To increase the comprehensiveness of the dataset library, we have collected \rawDataCnt widely used spatial-temporal datasets from 11 different countries covering various periods and processed them into the atomic files. We have also replicated \modelCnt classic spatial-temporal prediction models, including state-of-the-art models, that cover three categories and nine sub-categories of tasks: \emph{Macro Group Prediction Tasks} (\eg traffic flow prediction, traffic speed prediction, on-demand service prediction, traffic accident prediction, OD matrix prediction), \emph{Micro Individual Prediction Tasks} (\eg trajectory next-location prediction, travel time prediction), and \emph{Fundamental Tasks} (\eg map matching and road network representation learning). In addition, we have implemented rich auxiliary functions such as automatic parameter tuning and the visualization platform to facilitate the use of these datasets and models. We will continually incorporate more datasets and models into our library to provide more comprehensive benchmark tasks.
    \item \textbf{Diverse and Flexible Evaluation Metrics}: \name offers a range of standard evaluation metrics for assessing different types of spatial-temporal prediction models. These metrics cover typical tasks such as classification and regression, ensuring a comprehensive model performance evaluation. In addition, \name provides flexible evaluation strategies. For macro-level prediction tasks, data slicing and window settings are available to determine how the training, validation, and testing datasets are partitioned and the input/output data length for single-step and multi-step predictions. For micro-level prediction tasks, window settings enable the partitioning of trajectories to assess model performance on long, medium, and short trajectories. By combining the evaluation metric, dataset division, and window settings, users can conduct diverse and flexible evaluations of models belonging to the same task.
\end{itemize}

To the best of our knowledge, \name is the first open-source library for urban spatial-temporal prediction. We consider it an essential resource for exploring and developing spatial-temporal prediction models. By enabling fair model comparisons, designing a unified data storage format, and simplifying the process of developing new models, \name is poised to contribute to the spatial-temporal prediction field significantly. Additionally, \name helps to establish evaluation standards in the field and foster its fast-paced and standardized growth.

% The subsequent sections are organized as follows: Section~\ref{data} introduces the basic unit of spatial-temporal data and atomic file format, Section~\ref{tasks} introduces spatial-temporal data tasks and definitions, Section~\ref{model} introduces the development roadmap of models under different tasks, Section~\ref{library} introduces the \name open source library, Section~\ref{usage} introduces the use cases of \name, Section~\ref{experiments} introduces the model comparison experiments under different tasks based on \name, Section~\ref{cmp} introduces the comparison of \name with existing open source libraries, and Section~\ref{conclusion} provides a conclusion.

\section{RELATED WORK} \label{relate}
\subsection{Urban Spatial-temporal Prediction} 

Spatial-temporal prediction problems can be abstracted as a time series prediction problem, which is using a series of historical spatial-temporal factors to predict the spatial-temporal target. According to types of prediction targets, spatial-temporal prediction tasks can be subdivided into macro state prediction and microscopic micro individual prediction. The macro state state is a state that describes the spatial-temporal situation at the macro level, such as traffic flow, traffic speed, on-demand services, etc. The micro individual prediction is to predict the behavior of an individual user from a micro level, such as trajectory next-location prediction, travel time estimate, route planing, etc.

% \name mainly focuses on the \taskcnt widely researched spatial-temporal prediction tasks: Traffic Flow Prediction, Traffic Speed Prediction, On-demand Service Prediction, Traffic Accident Prediction, Origin-Destination Matrix Prediction, Trajectory next-location prediction, Estimated Time of Arrival, Map Matching and Road Network Representation Learning.

In general, regardless of the type of task, spatial-temporal prediction methods can be divided into two types: traditional methods and deep learning methods. Traditional methods are mainly based on statistical assumptions for sequence prediction, like Hidden Markov Model (HMM)~\cite{HMMM}, Support Vector Regression (SVR)~\cite{SVR}, and Matrix Factorization based Method FPMC-LR~\cite{FPMC}. However, these methods are shallow models and have limited ability to capture the nonlinearity of spatial-temporal data, therefore, these methods perform poorly in practical. Compared with traditional methods, deep learning methods have stronger feature learning capabilities and can automatically extract features from spatial-temporal data, which allows them to better capture spatio-temporal correlations. Thus, various deep leaning models such as convolutional neural network (CNN), recurrent neural network (RNN) and graph convolutional network (GCN) have achieved great success in spatial-temporal prediction field. Due to the powerful ability of RNN in modeling sequence data, RNN is often used to model the temporal correlation of spatial-temporal data in previous studies. In terms of spatial correlation modeling, CNN or GCN is often used. The traditional CNN models can only model Euclidean data, so researchers generally convert the spatial-temporal network structure at different times into images. In addition, the GCN models, including the spatial methods and the spectral methods, can directly model the spatial-temporal data of the graph structure and have achieved state-of-the-art results. Recently, spatial and temporal attention mechanisms have also been introduced into the field of spatial-temporal prediction to adaptively assign different importance to spatial-temporal data to capture dynamic temporal and spatial correlations.

% \subsection{Comparison with existing surveys}
\subsection{Related Libraries} To the best of our knowledge, \name~\cite{libcity} is the first spatial-temporal prediction library that enables researchers to conduct comprehensive comparative experiments and develop new models. Recently, other researchers have proposed benchmarks in spatial-temporal prediction similar to \name. DL-Traff~\cite{jiang2021dl} is an open-source project offering a traffic prediction benchmark using grid-based and graph-based models. DGCRN~\cite{DGCRN} summarizes previous work and produces a benchmark in traffic prediction. However, these two projects only accumulate the model codes from past research work without a modular design, which makes it inconvenient for users to use. Furthermore, they only focus on macro-level traffic prediction without contributing to micro-level individual prediction tasks. Microsoft FOST~\footnote{\url{https://github.com/microsoft/FOST}} (Forecasting Open Source Tool) is a general forecasting tool that aims to provide an easy-to-use tool for spatial-temporal forecasting, but its supported models and applications are limited. % PyTorch Geometric Temporal~\cite{rozemberczki2021pytorch} is a temporal (dynamic) extension library for PyTorch Geometric. This work only focuses on the application of spatial-temporal prediction models related to graph neural networks, ignoring others.

In addition, it is worth noting that numerous similar experimental libraries are available in other research fields. For instance, RecBole~\cite{recbole} is a recommendation algorithm framework that reproduces a vast range of recommendation models and provides various evaluation strategies and data preprocessing operations, making it easy to conduct experiments. Meanwhile, MMDetection~\cite{mmdetection} adopts a modular design, enabling researchers to efficiently develop new models based on it for object detection tasks. FastReID~\cite{he2020fastreid} continuously reproduces state-of-the-art models and releases corresponding pre-trained models for both research and industrial purposes.

\name combines the strengths of the libraries as mentioned above, such as various baseline models, diverse evaluation strategies, and modular design. As a result, it not only facilitates researchers to conduct experiments and develop new models but also promotes standardization within the spatial-temporal prediction field.

\section{Spatial-Temporal Data} \label{data}

\subsection{Urban Spatial-Temporal Data}
In modern cities, there are many urban information infrastructure devices such as Internet of Things (IoT) sensors, GPS terminals, smartphones, Location Based Services (LBS), Radio Frequency Identification (RFID), and wearable intelligent bracelets, which collect a lot of spatial-temporal big data related to the city.

The common urban spatial-temporal data contains three categories:

\begin{enumerate}
\item \textbf{Urban Scenes Data}: This kind of data is \textit{low-frequency}, \textit{static} urban structure data, such as urban map data, point-of-interest (POI) data, urban road network data, etc.
\item \textbf{Individual Behavior Data}: This kind of data is generally IoT data, which is \textit{high-frequency} and \textit{dynamic} spatial-temporal \textbf{trajectory} data, such as floating car GPS data, public transportation card data, cell phone signaling data, Location-Based Social Networks (LBSNs) data, etc.
\item \textbf{Group Dynamics Data}: This kind of data is generally the aggregated data of group behavior, which is \textit{high-frequency}, \textit{dynamic} spatial-temporal \textbf{attribute} data, such as population density, traffic road condition, ride demand, origin-destination (OD) network, climate, and weather data, etc.
\end{enumerate}

Urban spatial-temporal data is characterized by its dynamic changes over time and space. The distribution of primary urban points of interest and road networks are examples of spatial-temporal static data. On the other hand, urban traffic flow data and weather data recorded by IoT sensors are considered spatial static and temporal dynamic data. In contrast, user trajectory data and check-in data are examples of spatial-temporal dynamic data, which capture the movements and behaviors of individuals over time and space.

\subsection{Atomic files} \label{atomic_files}
Existing open-source spatial-temporal datasets are usually stored in different formats, such as NPZ, PKL, H5, CSV, etc, which invariably increases the difficulty and burden for users to use these datasets. Therefore, to provide a unified representation and store format of urban spatial-temporal data, we define five types of \textit{atomic files}, \ie five minimum information units of urban spatial-temporal data as follows:

\begin{itemize}
    \item \textbf{Geographical Unit Data}: Geographical Unit data are the basic units in spatial-temporal data, \ie the point, the line, and the plane.
    \item \textbf{User Unit Data}: User Unit data describes attribute information of desensitized urban activity participants.
    \item \textbf{Unit Relation Data}: Unit Relation data describes the relationships between units in urban spatial-temporal prediction scenarios.
    \item \textbf{Spatial-temporal Dynamic Data}: Spatial-temporal dynamic data describes the attribute information of entities in a city that dynamically changes over time, including  \textit{Spatial Static Temporal Dynamic (SSTD) Data} and \textit{Spatial Dynamic Temporal Dynamic (SDTD) Data}.
    \item \textbf{External Data}: External data describes the auxiliary information associated with urban geographical and user units.
\end{itemize}

For the above five types of atomic files, we use a comma-separated value format for data storage and define different file suffixes for different kinds of atomic files. In addition, we have restricted the information contained in each line in the atomic files. For example, the ".geo" file must contain ID, geographic entity type (point, line, polygon), and coordinate information. Other attributes must be stored after the above three columns, such as the POI category or the road width. More details can be found in Table~\ref{tab:atomic_files}. 

% The definition of spatial-temporal data storage format and spatial-temporal model input and output in \name has formed a team standard of Zhongguancun Smart City Industry Innovation Alliance~\cite{?}.
% 补充 该标准已经被...

\begin{table*}[t]
  \centering
  \caption{Summary of Atomic Files}
    \begin{tabular}{c|c}
    \toprule
    \textbf{Suffix} & \textbf{Content} \\
    \midrule
    .geo  & Geographical Unit Data \\
    .usr  & User Unit Data \\
    .rel  & Unit Relation Data \\
    .dyna & Spatial Dynamic Temporal Dynamic (SDTD) Data for User Unit (Trajectories) \\
    .dyna & Spatial Static Temporal Dynamic (SSTD) Data of Graph Network Unit Relation Data \\
    .grid & Spatial Dynamic Temporal Dynamic (SDTD) Data for Grid Relation Data \\
    .od   & Spatial Dynamic Temporal Dynamic (SDTD) Data for OD Relation Data \\
    .gridod & Spatial Dynamic Temporal Dynamic (SDTD) Data for Grid OD Relation Data \\
    .ext  & External Data \\
    \bottomrule
    \end{tabular}%
  \label{tab:atomic_files}%
\end{table*}%

\section{Spatial-Temporal Prediction Tasks} \label{tasks}

The primary purpose of urban spatial-temporal prediction tasks is to make forecasts based on historical observations for urban spatial-temporal data, including  \textit{Spatial Static Temporal Dynamic (SSTD) Data} and \textit{Spatial Dynamic Temporal Dynamic (SDTD) Data}. We classify the main spatial-temporal prediction tasks into two categories, one for \textbf{macro group prediction tasks} and one for \textbf{micro individual prediction tasks}. In addition to these two types of spatial-temporal prediction tasks in this study, some \textbf{fundamental tasks} that support urban spatial-temporal prediction are also considered. Specifically, the tasks targeted in this study are as follows:

\subsection{Macro Group Prediction Tasks} 
This type of task is mainly for the \textit{Spatial Static Temporal Dynamic (SSTD) Data}. Specially, these tasks are used to predict the macro group's spatial-temporal attributes in the future. Formally, given the \textit{SSTD} data, our goal is to learn a mapping function $f$ from the previous $T$ steps' observation value to predict future $T'$ steps' attributes~\cite{trafficsurvey1, trafficsurvey2},
\begin{equation}\label{eq:problem_def} \small
[\boldsymbol{X}_{(t-T+1)}, \cdots, \boldsymbol{X}_t; \mathcal{G}] \stackrel{f}{\longrightarrow} [{\boldsymbol{X}}_{(t+1)}, \cdots, {\boldsymbol{X}}_{(t+T')}].
\end{equation}
where the tensor $\boldsymbol{X}$ can be obtained from Graph Relation Data, Grid Relation Data, OD Relation data, and Grid OD Relation Data.

Typical macro group prediction tasks include traffic flow prediction, traffic speed prediction, on-demand service prediction, origin-destination matrix prediction, and traffic accidents prediction as follows:
\begin{itemize}
    \item \textit{Traffic Flow Prediction}~\cite{deepst, STResNet} aims to forecast the number of vehicles that will enter or exit specific regions or road segments in a future time window. Accurate predictions can significantly improve dynamic traffic control, route planning, navigation services, and other high-level applications.
    \item \textit{Traffic Speed Prediction}~\cite{DCRNN, STGCN} aims to forecast the average speed of vehicles over a specific road segment in the future. Here traffic speed refers to the distance traveled per unit of time, and the focus is on the average speed of vehicles across a particular road segment rather than on specific vehicles. This task is similar to traffic flow prediction, and both problems can be solved using similar prediction strategies.
    \item \textit{On-demand Service Prediction}~\cite{DMVSTNet, demand} aims to forecast the number of ride requests for a particular region or road segment. Short-term demand prediction is critical for on-demand ride-hailing platforms like Uber and Didi because dynamic pricing depends on real-time demand prediction, and dispatch systems can relocate drivers to high-demand areas. Typically, the number of pick-ups and drop-offs represents the demand in a specific region during a particular time interval.
    \item \textit{Origin-Destination Matrix Prediction}~\cite{OD1, OD2} aims to forecast transitions between different nodes, also known as edge flow or origin-destination-based flow. OD prediction provides more detailed insight into travel needs and enhances understanding of urban traffic patterns compared to other traffic prediction tasks.
    \item \textit{Traffic Accident Prediction}~\cite{accident1, accident2} is crucial for improving public safety. However, predicting the occurrence of traffic accidents in terms of their location and time is generally difficult. In recent studies, researchers have shifted their focus to forecasting the number of traffic accidents or the severity of traffic risks in specific regions. Such regions with the highest risk level can be considered hot spots, allowing for targeted safety measures to be implemented.
\end{itemize}

\subsection{Micro Individual Prediction Tasks} 
This type of task is mainly for the \textit{Spatial Dynamic Temporal Dynamic (SDTD) Data}. Specially, these tasks perform task-specific label prediction based on the user's historical trajectory data. Formally, given the \textit{SDTD} data, our goal is to learn a mapping function $f$ from user's historical trajectory $\mathcal{T}$ to predict a task-specific label~\cite{trajsurvey1, trajsurvey2},
\begin{equation}\label{eq:problem_def2} \small
\mathcal{T}=[\langle \boldsymbol{v}_i, t_i \rangle]_{i=1}^m \stackrel{f}{\longrightarrow} label.
\end{equation}
where the sample point $\boldsymbol{v}_i$ can be the Point, Line, or Plane Geographical Unit mentioned above. In other words, the trajectory $\mathcal{T}$ can be the GPS-based trajectories, POI-based trajectories, Road-network Constrained trajectories, and Region/Grid-based trajectories.

Typical micro individual prediction tasks include trajectory next-location prediction and travel time prediction (also called Estimated time of arrival, ETA) as follows:

\begin{itemize}
    \item \textit{Trajectory Next-Location Prediction}~\cite{STRNN, DeepMove} aims to predict the location that a user may visit next, given the historical trajectory of this user. Formally, given the historical trajectory $\mathcal{T}=[\langle \boldsymbol{v}_i, t_i \rangle]_{i=1}^m$ of user $i$, our goal is to estimate the probability of location $\boldsymbol{v}_{m+1}$ for user $i$ in the next timestamp $t_{m+1}$. Most of the next-location prediction research focuses on the POI-based trajectories.
    \item \textit{Travel Time Prediction}~\cite{eta1, eta2} aims to predict the arrival time given the trajectory and departure time. Given the trajectory sequence $\{\boldsymbol{v}_1,\dots,\boldsymbol{v}_m\}$ and the departure time $t_1$, our goal is to estimate the travel time between the starting point $\boldsymbol{v}_1$ and the destination $\boldsymbol{v}_m$ through $\{\boldsymbol{v}_1,\dots,\boldsymbol{v}_m\}$. Besides, in some ETA scenarios, only the starting point, departure time, and destination are given without the specific trajectory sequence. Most of the travel time prediction research focuses on the GPS-based trajectories, Road-network Constrained trajectories, and Region or Grid-based trajectories.
\end{itemize}

\subsection{Fundamental tasks} 
Fundamental tasks provide support for macro and micro prediction tasks mentioned above. The fundamental tasks considered in this work include map matching and road network representation learning.
\begin{itemize}
    \item \textit{Map Matching}~\cite{mapmatch1, mapmatch2} aims to match the raw GPS trajectories to the road segments of the road network. Given a GPS-based trajectory $\mathcal{T}^{gps}$ and a road network $\mathcal{G} = (\mathcal{V}, \mathcal{E}, \boldsymbol{F}_\mathcal{V}, \boldsymbol{A})$, our goal is to find a road-network constrained trajectory $\mathcal{T}$ that matches $\mathcal{T}^{gps}$ with its real path.
    \item \textit{Road Network Representation Learning}~\cite{RFN, IRN2Vec} aims to discover the representations from raw road network data, similar to network embedding methods\cite{wang2017community}. Given a road network $\mathcal{G} = (\mathcal{V}, \mathcal{E}, \boldsymbol{F}_\mathcal{V}, \boldsymbol{A})$, where $\boldsymbol{F}_\mathcal{V} \in \mathbb{R}^{N \times D}$ contains raw attributes, our goal is to derive a $d$-dimensional representation $\boldsymbol{F}_\mathcal{V}^{dense} \in \mathbb{R}^{N \times d}$ for each road segment on $\mathcal{G}$, where $d \ll D$. $\boldsymbol{F}_\mathcal{V}^{dense}$ is supposed to preserve the multifaceted characteristics of $\boldsymbol{F}_\mathcal{V}$.
\end{itemize}

\section{The Library: \name} \label{library}
\begin{figure}[t]
    \centering
    \includegraphics[width=0.95\columnwidth]{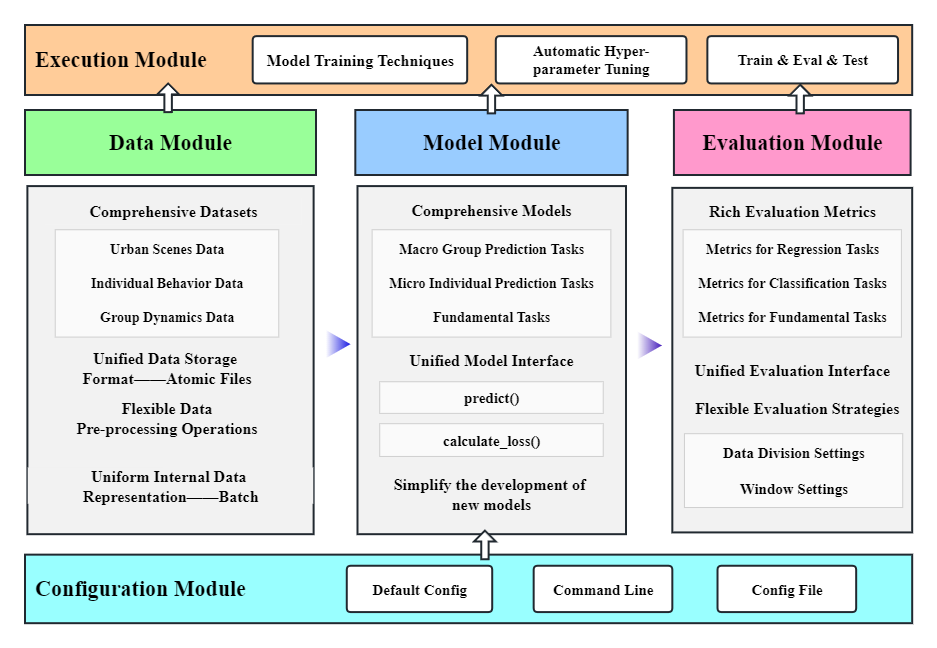}
    \caption{Overview of the \name library}
    \vspace{-0.3cm}
    \label{fig:framework}
\end{figure}

Deep learning algorithms have been extensively used in urban spatial-temporal prediction in recent years, resulting in a wealth of research findings. However, according to our survey, less than 30\% of the papers published in 11 leading conferences and journals have made their code and data open source, which hampers reproducibility in the field. Additionally, the need for recognized state-of-the-art (SOTA) models, standardized datasets, and transparent experimental settings creates obstacles to assessing the performance of new models and ultimately stifles innovation. In contrast, domains like Computer Vision, Natural Language Processing, and Recommendation Systems have standardized datasets such as IMAGENET~\cite{IMAGENET} and algorithm libraries like MMDetection~\cite{mmdetection}, and RecBole~\cite{recbole}. Unfortunately, urban spatial-temporal prediction lacks these resources. We propose LibCity, a comprehensive and unified library for urban spatial-temporal prediction, to address this issue.

Figure~\ref{fig:framework} illustrates the framework of \name, which comprises five main modules, including the data, model, evaluation, execution, and configuration module. Combining these modules forms a cohesive pipeline that provides researchers with a reliable experimental environment, and each module is responsible for a specific step in the pipeline. The subsequent sections will provide a detailed description of the implementation of each module.

\begin{itemize}
\item \textbf{Data Module}: Responsible for loading datasets and data preprocessing.
\item \textbf{Model Module}: Responsible for initializing the reproduced baseline model or custom model.
\item \textbf{Evaluation Module}: Responsible for evaluating model prediction results through multiple metrics.
\item \textbf{Execution Module}: Responsible for model training and prediction.
\item \textbf{Configuration Module}: Responsible for managing all the parameters involved in the framework.
% \item \textbf{Utils Module}: Responsible for providing users with practical tools.
\end{itemize}

\subsection{Data Module}

\begin{figure}[t]
    \centering
    \includegraphics[width=0.95\columnwidth]{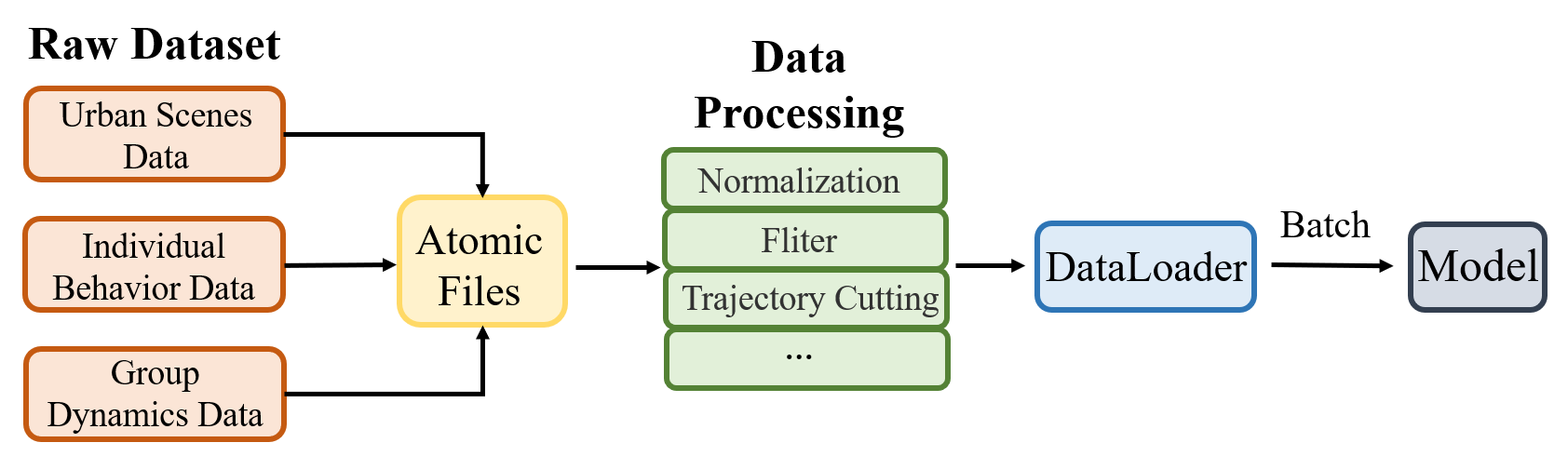}
    \caption{The data processing flow in \name}
    \label{fig:dataflow}
    \vspace{-0.4cm}
\end{figure}

Our data module aims to address the issue of unfair evaluation caused by different data preparation methods by establishing a standardized data processing flow, as shown in Figure~\ref{fig:dataflow}. This flow encompasses two types of data: user-oriented and model-oriented. The former establishes a unified storage format for spatial-temporal data, referred to as \textit{atomic files}, which provides users with a consistent data input format. The latter defines a key-value data structure, known as \textit{Batch}, to facilitate uniform data interaction between the data module and model module. Once the atomic files have been loaded and preprocessed, the \textit{Dataloader} class in PyTorch is employed to convert the data into a \textit{Batch} structure and then fed to the model module.

\subsubsection{Atomic Files} \name defines a general and extensible data storage format for urban spatial-temporal data, \ie \textit{atomic files}. Atomic files contain five categories, the minor information units in spatial-temporal data. For more details, please refer to Section~\ref{atomic_files} above.

\subsubsection{Preprocessing Operations} The \name library provides support for various data preprocessing operations. For the macro group prediction task, it is essential to normalize the spatial-temporal data to improve the model's convergence toward the optimal solution. \name supports multiple data normalization methods, including Z-score normalization, min-max normalization, logarithmic normalization, and other custom normalization methods, which can be easily achieved through parameter settings. \name also supports the handling of missing values and outliers. For the micro individual prediction task,  \name incorporates two trajectory filtering methods: inactive user filtering and inactive POI filtering. Inactive users can be filtered out by evaluating their activity based on the number and length of their trajectories and setting the minimum number of trajectories and minimum trajectory length. Similarly, inactive POIs can be filtered out based on the minimum number of visits. In this way, we can filter the data to reduce the impact of sparsity in spatial-temporal data. 

\subsubsection{Batch} The \textit{Batch} is a key-value data structure based on the implementation of \textit{python.dict}. It consists of feature names as keys and corresponding feature tensors (\textit{torch.Tensor}) in a mini-batch as values. The purpose of the \textit{Batch} is to facilitate data interaction between the data module and the model module in a systematic pipeline. In the pipeline, the executor extracts one \textit{Batch} object from the \textit{Dataloader} at a time and feeds it into the model. This structure enables the convenient use of feature tensors by referring to corresponding feature names. As different prediction models use varying features, they can all be stored in the form of \textit{Batch}. Through this data form, \name can construct unified model interfaces and implement general executor for model training and testing, which is particularly useful for developing new models.

\subsubsection{Comprehensive Datasets} We have conducted a comprehensive literature survey on spatial-temporal prediction and selected \paperCnt representative papers, including survey papers. We identified all the open datasets used from these papers and selected \rawDataCnt datasets based on their popularity, time span length, and data size. These datasets cover all the \taskcnt tasks that \name supports, consisting of 40 group dynamics datasets (\textit{SSTD data}), ten individual behavior datasets (\textit{SDTD data}), and five urban scenes datasets (\textit{road network data}). Table~\ref{tab:tsdataset}, Table~\ref{tab:trajdataset}, and Table~\ref{tab:rndataset} provide statistics on these datasets. To facilitate the use of these datasets in \name, we have converted all of them into the atomic file format and created conversion tools, which can be found at this link\footnote{\url{https://github.com/LibCity/Bigscity-LibCity-Datasets}}.

% Table generated by Excel2LaTeX from sheet '数据集总结'
\begin{table*}[hbtp]
  \centering
  \caption{Summary of Group Dynamics Datasets (\textit{SSTD Data}) in \name}
  \resizebox{\textwidth}{!}{
    \begin{tabular}{cccccccp{9em}}
    \toprule
    {\textbf{DATASET}} & 
    {\textbf{\#GEO}} & 
    {\textbf{\#REL}} & 
    {\textbf{\#DYNA}} & 
    {\textbf{PLACE}} & 
    {\textbf{DURATION}} & 
    {\textbf{\#TS}} & 
    \textbf{DATA TYPE} \\
    \midrule
    METR-LA~\cite{DCRNN} & 207   & 11,753  & 7,094,304  & Los Angeles, USA & Mar. 1, 2012 - Jun. 27, 2012 & 5min  & Graph Speed \\
    \midrule
    Los-Loop~\cite{TGCN} & 207   & 42,849  & 7,094,304  & Los Angeles, USA & Mar. 1, 2012 - Jun. 27, 2012 & 5min  & Graph Speed \\
    \midrule
    SZ-Taxi~\cite{TGCN} & 156   & 24,336  & 464,256  & Shenzhen, China & Jan. 1, 2015 - Jan. 31, 2015 & 15min & Graph Speed \\
    \midrule
    Q-Traffic~\cite{bbliaojqZhangKDD18deep} & 45,148  & 63,422  & 264,386,688  & Beijing, China & Apr. 1, 2017 - May 31, 2017 & 15min & Graph Speed \\
    \midrule
    Loop Seattle~\cite{cui2018deep, TGCLSTM} & 323   & 104,329  & 33,953,760  & Greater Seattle Area, USA & Jan. 1, 2015 - Dec. 31, 2015 & 5min  & Graph Speed \\
    \midrule
    PEMSD7(M)~\cite{STGCN} & 228   & 51,984  & 2,889,216  & California, USA & Weekdays of May. Jun., 2012 & 5min  & Graph Speed \\
    \midrule
    PEMS-BAY~\cite{DCRNN} & 325   & 8,358  & 16,937,700  & San Francisco Bay Area, USA & Jan. 1, 2017 - Jun. 30, 2017 & 5min  & Graph Speed \\
    \midrule
    Rotterdam~\cite{guopeng2020dynamic}  & 208   & - & 4,813,536  & Rotterdam, Holland & 135 days of 2018 & 2min  & Graph Speed \\
    \midrule
    PeMSD3~\cite{STSGCN} & 358   & 547   & 9,382,464  & California, USA & Sept. 1, 2018 - Nov. 30, 2018 & 5min  & Graph Flow \\
    \midrule
    PEMSD7~\cite{STSGCN} & 883   & 866   & 24,921,792  & California, USA & Jul. 1, 2016 - Aug. 31, 2016 & 5min  & Graph Flow \\
    \midrule
    Beijing subway~\cite{ResLSTM} & 276   & 76,176  & 248,400  & Beijing, China & Feb. 29, 2016 - Apr. 3, 2016 & 30min & Graph Flow \\
    \midrule
    M-dense~\cite{de2020spatio} & 30    & - & 525,600  & Madrid, Spain & Jan. 1, 2018 - Dec. 21, 2019 & 60min & Graph Flow \\
    \midrule
    SHMetro~\cite{liu2020physical} & 288   & 82,944  & 1,934,208  & Shanghai, China & Jul. 1, 2016 - Sept. 30, 2016 & 15min & Graph Flow \\
    \midrule
    HZMetro~\cite{liu2020physical} & 80    & 6,400  & 146,000  & Hangzhou, China & Jan. 1, 2019 - Jan. 25, 2019 & 15min & Graph Flow \\
    \midrule
    NYCTaxi-Dyna\footnotemark[3] & 263   & 69,169  & 574,392  & New York, USA & Jan. 1, 2020 - Mar. 31, 2020 & 60min & Region Flow \\
    \midrule
    PeMSD4~\cite{ASTGCN} & 307   & 340   & 5,216,544  & San Francisco Bay Area, USA & Jan. 1, 2018 - Feb. 28, 2018 & 5min  & Graph Flow, Speed, Occupancy \\
    \midrule
    PEMSD8~\cite{ASTGCN} & 170   & 277   & 3,035,520  & San Bernardino Area, USA & Jul. 1, 2016 - Aug. 31, 2016 & 5min  & Graph Flow, Speed, Occupancy \\
    \midrule
    TaxiBJ2013~\cite{STResNet} & {32*32} & - & 4,964,352  & Beijing, China & Jul. 1, 2013 - Oct. 30, 2013 & 30min & Grid In\&Out Flow \\
    \midrule
    TaxiBJ2014~\cite{STResNet} & {32*32} & - & 4,472,832  & Beijing, China & Mar. 1, 2014 - Jun. 30, 2014 & 30min & Grid In\&Out Flow \\
    \midrule
    TaxiBJ2015~\cite{STResNet} & {32*32} & - & 5,652,480  & Beijing, China & Mar. 1, 2015 - Jun. 30, 2015 & 30min & Grid In\&Out Flow \\
    \midrule
    TaxiBJ2016~\cite{STResNet} & {32*32} & - & 6,782,976  & Beijing, China & Nov. 1, 2015 - Apr. 10, 2016 & 30min & Grid In\&Out Flow \\
    \midrule
    T-Drive~\cite{yuan2011driving, yuan2010t} & {32*32} & -  & 3,686,400  & Beijing, China & Feb. 1, 2015 - Jun. 30, 2015 & 60min & Grid In\&Out Flow \\
    \midrule
    Porto\footnotemark[3] & {20*10} & - & 441,600  & Porto, Portugal & Jul. 1, 2013 - Sept. 30, 2013 & 60min & Grid In\&Out Flow \\
    \midrule
    NYCTaxi140103\footnotemark[4] & {10*20} & - & 432,000  & New York, USA & Jan. 1, 2014 - Mar. 31, 2014 & 60min & Grid In\&Out Flow \\
    \midrule
    NYCTaxi140112~\cite{ACFM} & {15*5} & - & 1,314,000  & New York, USA & Jan. 1, 2014 - Dec. 31, 2014 & 30min & Grid In\&Out Flow \\
    \midrule
    NYCTaxi150103~\cite{STDN} & {10*20} & - & 576,000  & New York, USA & Jan. 1, 2015 - Mar. 1, 2015 & 30min & Grid In\&Out Flow \\
    \midrule
    NYCTaxi160102~\cite{DSAN} & {16*12} & - & 552,960  & New York, USA & Jan. 1, 2016 - Feb. 29, 2016 & 30min & Grid In\&Out Flow \\
    \midrule
    NYCBike140409~\cite{STResNet} & {16*8} & - & 562,176  & New York, USA & Apr. 1, 2014 - Sept. 30, 2014 & 60min & Grid In\&Out Flow \\
    \midrule
    NYCBike160708~\cite{STDN} & {10*20} & - & 576,000  & New York, USA & Jul. 1, 2016 - Aug. 29, 2016 & 30min & Grid In\&Out Flow \\
    \midrule
    NYCBike160809~\cite{DSAN} & {14*8} & - & 322,560  & New York, USA & Aug. 1, 2016 - Sept. 29, 2016 & 30min & Grid In\&Out Flow \\
    \midrule
    NYCBike200709\footnotemark[5] & {10*20} & - & 441,600  & New York, USA & Jul. 1, 2020 - Sept. 30, 2020 & 60min & Grid In\&Out Flow \\
    \midrule
    AustinRide\footnotemark[6] & {16*8} & - & 282,624  & Austin, USA & Jul. 1, 2016 - Sept. 30, 2016 & 60min & Grid In\&Out Flow \\
    \midrule
    BikeDC\footnotemark[7] & {16*8} & - & 282,624  & Washington, USA & Jul. 1, 2020 - Sept. 30, 2020 & 60min & Grid In\&Out Flow \\
    \midrule
    BikeCHI\footnotemark[8] & {15*18} & - & 596,160  & Chicago, USA & Jul. 1, 2020 - Sept. 30, 2020 & 60min & Grid In\&Out Flow \\
    \midrule
    NYCTaxi-OD\footnotemark[4] & 263   & 69,169  & 150,995,927  & New York, USA & Apr. 1, 2020 - Jun. 30, 2020 & 60min & OD Flow \\
    \midrule
    NYC-TOD~\cite{CSTN} & {15*5} & - & 98,550,000  & New York, USA & Jan. 1, 2014 - Dec. 31, 2014     & 30min & Grid-OD Flow \\
    \midrule
    NYCTaxi150103~\cite{STDN} & {10*20} & - & 115,200,000  & New York, USA & Jan. 1, 2015 - Mar. 1, 2015 & 30min & Grid-OD Flow \\
    \midrule
    NYCBike160708~\cite{STDN} & {10*20} & - & 115,200,000  & New York, USA & Jul. 1, 2016 - Aug. 29, 2016 & 30min & Grid-OD Flow \\
    \midrule
    NYC-Risk~\cite{GSNet} & 243   & 59,049  & 3,504,000  & New York, USA & Jan. 1, 2013 - Dec. 31, 2013 & 60min & Risk \\
    \midrule
    CHI-Risk~\cite{GSNet} & 243   & 59,049  & 3,504,000  & New York, USA & Jan. 1, 2013 - Dec. 31, 2013 & 60min & Risk \\
    \bottomrule
    \end{tabular}
    }
  \label{tab:tsdataset}%
\end{table*}%
\footnotetext[3]{\url{https://www.kaggle.com/c/pkdd-15-predict-taxi-service-trajectory-i}}
\footnotetext[4]{\url{https://www1.nyc.gov/site/tlc/about/tlc-trip-record-data.page}}
\footnotetext[5]{\url{https://www.citibikenyc.com/system-data}}
\footnotetext[6]{\url{https://data.world/ride-austin/ride-austin-june-6-april-13}}
\footnotetext[7]{\url{https://www.capitalbikeshare.com/system-data}}
\footnotetext[8]{\url{https://www.divvybikes.com/system-data}}
\footnotetext[9]{\url{https://www.openstreetmap.org}}

% Table generated by Excel2LaTeX from sheet '数据集总结'
\begin{table*}[t]
  \centering
  \caption{Summary of Individual Behavior Datasets (\textit{SDTD Data}) in \name}
  \resizebox{\textwidth}{!}{
    \begin{tabular}{cccccccc}
    \toprule
    {\textbf{DATASET}} & {\textbf{\#GEO}} & \textbf{\#REL} & {\textbf{\#USR}} & {\textbf{\#DYNA}} & \textbf{PLACE} & \textbf{DURATION} & \textbf{DATA TYPE} \\
    \midrule
    Seattle~\cite{HMMM} & 613,645  & {857,406 } & 1     & 7,531  & Seattle WA, USA & Jan. 17, 2009 & GPS-based \\
    \midrule
    Global~\cite{global} & 11,045  & {18,196 } & 1     & 2,502  & 100 cities & -     & GPS-based \\
    \midrule
    CD-Taxi-Sample~\cite{DeepTTE} & - & -     & 4,565  & 712,360  & Chengdu, China & Aug. 3, 2014 - Aug. 30, 2014 & GPS-based \\
    \midrule
    BJ-Taxi-Sample~\cite{DeepTTE} & 16,384  & -     & 76    & 518,424  & Beijing, China & Oct. 1, 2013 - Oct. 31, 2013 & GPS-based \\
    \midrule
    Porto~\cite{START} & 10,903  & {26,161 } & 435   & 695,085  & Porto, Portugal & Jul. 1, 2013 - Jul. 1, 2014 & Road-network Constrained \\
    \midrule
    Foursquare-TKY~\cite{10.1145/2814575} & 61,857  & -    & 2,292  & 573,703  & Tokyo, Japan & Apr. 4, 2012 - Feb. 16, 2013 & POI-based \\
    \midrule
    Foursquare-NYC~\cite{10.1145/2814575} & 38,332  & -     & 1,082  & 227,428  & New York, USA & Apr. 3, 2012 - Feb. 15, 2013 & POI-based \\
    \midrule
    Gowalla~\cite{10.1145/2020408.2020579} & 1,280,969  & {913,660 } & 107,092  & 6,442,892  & Global & Feb. 4, 2009 - Oct. 23, 2010 & POI-based \\
    \midrule
    BrightKite~\cite{10.1145/2020408.2020579} & 772,966  & {394,334 } & 51,406  & 4,747,287  & Global & Mar. 21, 2008 - Oct. 18, 2010 & POI-based \\
    \midrule
    Instagram~\cite{ijcai2018-458} & 13,187  & -     & 78,233  & 2,205,794  & New York, USA & Jun. 15, 2011 - Nov. 8, 2016 & POI-based \\
    \bottomrule
    \end{tabular}%
    }
  \label{tab:trajdataset}%
\end{table*}%

% Table generated by Excel2LaTeX from sheet '数据集总结'
\begin{table}[t]
\small
  \centering
  \caption{Summary of Urban Scenes Datasets (Road Network Data) in \name}
  \resizebox{\columnwidth}{!}{
    \begin{tabular}{cccc}
    \toprule
    {\textbf{DATASET}} & {\textbf{\#GEO}} & {\textbf{\#REL}} & \textbf{PLACE} \\
    \midrule
    BJ-Roadmap-Edge~\cite{START} & 40,306  & 101,024  & Beijing, China \\
    \midrule
    BJ-Roadmap-Node\footnotemark[9] & 16,927  & 38,027  & Beijing, China \\
    \midrule
    CD-Roadmap-Edge\footnotemark[9] & 6,195  & 15,962  & Chengdu, China \\
    \midrule
    XA-Roadmap-Edge\footnotemark[9] & 5,269  & 13,032  & Xian, China \\
    \midrule
    Porto-Roadmap-Edge~\cite{START} & 11,095  & 26,161  & Porto, Portugal \\
    \bottomrule
    \end{tabular}%
    }
  \label{tab:rndataset}%
\end{table}%

\subsection{Model Module}
To increase the modularity of the library and reduce coupling between different modules, \name uses a separate model module to implement classic spatial-temporal prediction algorithms. This module contains various models such as LSTM, GRU, TCN, GCN, etc. Encapsulating each model in a separate class, \name enables users to easily switch between different models and extend the library with new ones.

\subsubsection{Unified Interface} In specific, \name provides two standard interfaces for all urban spatial-temporal prediction models: \textit{predict()} and \textit{calculate\_loss()} functions as follows:

\begin{itemize}
    \item The \textit{predict()} function is used in the process of model prediction to return the model prediction results.
    \item The \textit{calculate\_loss()} function is used in the model training process to return the loss value, which needs to be optimized. 
\end{itemize}

Both methods take the internal data representation \textit{Batch} as input. These interface functions are general to different spatial-temporal prediction models, which allows researchers to implement various models in a highly unified way. When developing a new model, researchers only need to instantiate these two interfaces to connect with other modules in \name. They do not need to worry about how each of the other parts works. This design simplifies the development process and accelerates the development of new models.

\subsubsection{Implemented Models} Currently, \name supports \taskcnt mainstream spatial-temporal prediction tasks. Through careful investigation of the development process in the field of spatial-temporal prediction, we have selected \modelCnt classic spatial-temporal prediction models to reproduce, ranging from early CNN-based models to recent GCN-based models and hybrid models. Moreover, in order to cover a wide range of prediction models, we also implement four shallow baseline models. We have tested all implemented models' performance on at least two datasets. We summarize the implemented \modelCnt models in Table~\ref{tab:models}. Referring to Section~\ref{macro_models}, we also provide in the table the different basic structures of the model in the spatial and temporal dimensions. Users can refer to this table to learn about the main techniques and developments in the field of urban spatial-temporal prediction. % For papers that provide their original dataset, the performance of our reproduced model on this dataset is generally similar to that reported in the original paper.

% table* generated by Excel2LaTeX from sheet 'Sheet1'
\begin{table*}[htbp]
  \centering
  \caption{Implement Models in \name}
  \vspace{-0.3cm}
  \resizebox{\textwidth}{!}{
    \begin{tabular}{c|c|c|c|ccc|cccc}
    \toprule
    \multirow{2}[2]{*}{Task} & \multirow{2}[2]{*}{Model} & \multirow{2}[2]{*}{Conference} & \multirow{2}[2]{*}{Year} & \multicolumn{3}{c|}{Spatial Axis} & \multicolumn{4}{c}{Temporal axis} \\
          &       &       &       & CNN   & GCN   & Attn. & LSTM  & GRU   & TCN   & Attn. \\
    \midrule
    \multirow{3}[2]{*}{{Traditional Methods}} 
    & HA    &    -   &   -    &    -   &   -    &     -  &    -   &    -   &    -   & - \\
    & SVR~\cite{SVR}   &   NIPS    &   1996    &    -   &    -   &   -    &    -   &     -  &     - &  - \\
    & ARIMA~\cite{ARIMA} &   J TRANSP ENG
    &   2003    &    -   &    -   &   -    &    -   &     -  &     - &  - \\
    & VAR~\cite{VAR}   &   Princeton Press    &   1994    &    -   &    -   &   -    &    -   &     -  &     - &  - \\
    \midrule
    \multirow{4}[2]{*}{General Macro Group Prediction} & RNN~\cite{Seq2Seq}   &    NIPS   &   2014    &       &       &       & \checkmark     & \checkmark     &       &  \\
          & Seq2Seq~\cite{Seq2Seq} & NIPS  & 2014  &       &       &       & \checkmark     & \checkmark     &       &  \\
          & AutoEncoder~\cite{AutoEncoder} & IEEE TITS & 2014  &    -   &    -   &   -    &    -   &     -  &    -   & - \\
          & FNN~\cite{DCRNN} & ICLR & 2018  &    -   &    -   &   -    &    -   &     -  &    -   & - \\
    \midrule
    \multirow{16}[2]{*}{Traffic Flow Prediction} 
          & ST-ResNet~\cite{STResNet} & AAAI  & 2017  & \checkmark     &       &       &       &       &       &  \\
          & {STNN~\cite{STNN}} & ICDM  & 2017  &    -   &    -   &   -    &    -   &     -  &    -   & - \\
          & ACFM~\cite{ACFM} & ACM MM & 2018  & \checkmark     &       &     \checkmark  & \checkmark     &       &       &  \\
          & {STDN~\cite{STDN}} & AAAI  & 2019  &   \checkmark    &       &       &   \checkmark    &       &       & \checkmark \\
          & ASTGCN~\cite{ASTGCN} & AAAI  & 2019  &       & \checkmark     & \checkmark     &       &       &   \checkmark   & \checkmark \\
          & MSTGCN~\cite{ASTGCN} & AAAI  & 2019  &       & \checkmark     &       &       &       &   \checkmark     &  \\
          & {DSAN~\cite{DSAN}} & KDD   & 2020  &       &       &   \checkmark    &       &       &       &  \checkmark \\
          & STSGCN~\cite{STSGCN} & AAAI  & 2020  &       &   \checkmark    &       &      &       &       &  \\
          & AGCRN~\cite{AGCRN} & NIPS  & 2020  &       & \checkmark     &       &       & \checkmark     &       &  \\
          & CRANN~\cite{CRANN} & arXiv  & 2020  &    \checkmark   &       &   \checkmark    &    \checkmark   &       &       &  \\
          & {CONVGCN~\cite{CONVGCN}} & IET ITS & 2020  &   \checkmark    &   \checkmark    &       &       &       &       &  \\
          & {ResLSTM~\cite{ResLSTM}} & IEEE TITS & 2020  &       &     \checkmark  &       &    \checkmark   &       &       &  \checkmark \\
          & {MultiSTGCnet~\cite{MultiSTGCnet}} & IJCNN & 2020  &       &    \checkmark   &       &   \checkmark    &       &       &  \\
          & {ToGCN~\cite{ToGCN}} & IEEE TITS & 2021  &       & \checkmark     &       & \checkmark     &       &       &  \\
          & DGCN~\cite{DGCN} & IEEE TITS & 2022  &       &   \checkmark    &   \checkmark    &    \checkmark   &       &   \checkmark    & \checkmark \\
    \midrule
    \multirow{13}[2]{*}{Traffic Speed Prediction} & DCRNN~\cite{DCRNN} & ICLR  & 2018  &       & \checkmark     &       &       & \checkmark     &       &  \\
          & STGCN~\cite{STGCN} & IJCAI & 2018  &       & \checkmark     &       &       &       & \checkmark     &  \\
          & GWNET~\cite{GWNET} & IJCAI & 2019  &       & \checkmark     &       &       &       & \checkmark     &  \\
          & TGCN~\cite{TGCN} & IEEE TITS & 2019  &       & \checkmark     &       &       & \checkmark     &       &  \\
          & TGCLSTM~\cite{TGCLSTM} & IEEE TITS & 2019  &       & \checkmark     &       & \checkmark     &       &       &  \\
          & MTGNN~\cite{MTGNN} & KDD   & 2020  &       & \checkmark     &       &       &       & \checkmark     &  \\
          & {GMAN~\cite{GMAN}} & AAAI  & 2020  &      &       &    \checkmark   &       &       &       &  \checkmark \\
          & {ATDM~\cite{ATDM}} & NCA & 2021  &       &    \checkmark   &       &   \checkmark    &       &       &  \\
          & {STAGGCN~\cite{STAGGCN}} & CIKM  & 2020  &       &   \checkmark    &   \checkmark    &       &       &   \checkmark    &  \checkmark \\
          & {ST-MGAT~\cite{STMGAT}} & ICTAI  & 2020  &       &      &    \checkmark   &       &       &  \checkmark   &   \\
          & {DKFN~\cite{DKFN}} & ACM GIS & 2020  &       & \checkmark     &       & \checkmark     &       &       &  \\
          & {STTN~\cite{STTN}} & arXiv  & 2020  &       &  \checkmark     &    \checkmark   &       &       &     &  \checkmark  \\
          & {HGCN~\cite{HGCN}} & AAAI  & 2021  &       &    \checkmark   &       &       &       &    \checkmark    &  \checkmark  \\
          & GTS~\cite{GTS} &    ICLR   &   2021    &       & \checkmark     &       &       & \checkmark     &       &  \\
    \midrule
    \multirow{3}[2]{*}{On-Demand Service Prediction} 
    & {DMVSTNet~\cite{DMVSTNet}} &   AAAA   & 2018      &   \checkmark    &       &       &    \checkmark   &      &       &  \\
    & {STG2Seq~\cite{STG2Seq}} &   IJCAI    &  2019     &      &   \checkmark    &       &       &       &       & \checkmark \\
    & {CCRNN~\cite{CCRNN}} &    AAAI   &   2021   &      &   \checkmark    &       &       &   \checkmark    &       &  \\
    \midrule
    \multirow{1}[2]{*}{Traffic Accident Prediction} & {GSNet~\cite{GSNet}} & AAAI & 2021  &       &   \checkmark    &       &       &    \checkmark   &       &  \checkmark \\
    \midrule
    \multirow{2}[2]{*}{OD Matrix Prediction} & {GEML~\cite{GEML}} & KDD & 2019  &       &   \checkmark    &       &   \checkmark    &       &       &  \\
          & {CSTN~\cite{CSTN}} & IEEE TITS & 2019  &  \checkmark      &       &       & \checkmark       &       &       &  \\
    \midrule
    \multirow{11}[2]{*}{Trajectory Next-Location Prediction} & FPMC~\cite{FPMC} & WWW   & 2010  &       &       &       &       &       &       &  \\
          & LSTM~\cite{RNN2} &  CoRR     &   2013    &       &       &       &  \checkmark      &       &       &  \\
          & ST-RNN~\cite{STRNN} & AAAI  & 2016  &       &       &       &       &  \checkmark      &       &  \\
          & SERM~\cite{SERM} & CIKM  & 2017  &       &       &       &       &  \checkmark     &       &  \\
          & DeepMove~\cite{DeepMove} & WWW   & 2017  &       &       &       &  \checkmark    &       &     & \checkmark \\
          & {CARA~\cite{CARA}} & SIGIR & 2018  &       &       &       &       & \checkmark      &       &  \\
          & {HSTLSTM~\cite{HSTLSTM}} & IJCAI & 2018  &       &       &       &  \checkmark     &       &       &  \\
          & ATSTLSTM~\cite{ATSTLSTM} & IEEE TSC  & 2019  &       &       &       &  \checkmark     &       &       & \checkmark \\
          & LSTPM~\cite{LSTPM} & AAAI  & 2020  &       &       &       &       &  \checkmark     &       & \checkmark \\
          & {GeoSAN~\cite{GeoSAN}} & KDD   & 2020  &       &       &       &       &       &       & \checkmark \\
          & {STAN~\cite{STAN}} & WWW   & 2021  &       &       &       &       &       &       & \checkmark \\
    \midrule
    \multirow{2}[2]{*}{Travel Time Prediction} & {DeepTTE~\cite{DeepTTE}} & AAAI  & 2018  & \checkmark      &       &       & \checkmark      &       &       &  \\
          & TTPNet~\cite{TTPNet} & TKDE  & 2022  & \checkmark      &       &       &  \checkmark     &       &       &  \\
    \midrule
    \multirow{3}[2]{*}{Map Matching} & STMatching~\cite{STMatching} & ACM GIS & 2009  &    -   &    -   &     -  &    -   &     -  &   -    &  - \\
    & HMMM~\cite{HMMM} & ACM GIS & 2009  &    -   &    -   &     -  &    -   &     -  &   -    &  - \\
    & IVMM~\cite{IVMM} & IEEE MDM & 2010 &    -   &    -   &     -  &    -   &     -  &   -    &  - \\
    \midrule
    \multirow{6}[2]{*}{Road Network Representation Learning}
    & DeepWalk~\cite{DeepWalk} & KDD & 2014   &    -   &    -   &     -  &    -   &     -  &   -    &  - \\
    & LINE~\cite{LINE} & WWW   & 2015  &    -   &    -   &     -  &    -   &     -  &   -    &  - \\
    & Node2Vec~\cite{node2vec} & KDD & 2016   &    -   &    -   &     -  &    -   &     -  &   -    &  - \\
    & ChebConv~\cite{ChebConv} & NIPS  & 2016  &       &   \checkmark    &       &       &       &       &  \\
    & GAT~\cite{GAT} & arXiv & 2017  &       &      &   \checkmark    &       &       &       &  \\
    & GeomGCN~\cite{GeomGCN} & ICLR  & 2020  &       &   \checkmark    &       &       &       &       &  \\
    \bottomrule
    \end{tabular}%
    }
  \label{tab:models}%
\end{table*}%

% Table generated by Excel2LaTeX from sheet '模型总结'
% \begin{table*}[htbp]
%   \centering
%   \caption{Summary of Implemented Models in \name}
%     \begin{tabular}{c|p{6.7em}|p{6em}|p{10em}|p{12em}}
%     \toprule
%     \textbf{Task} & \multicolumn{1}{p{6.7em}<{\centering}|}{Traditional} & \multicolumn{1}{p{6em}<{\centering}|}{CNN-based} & \multicolumn{1}{p{10em}<{\centering}|}{RNN-based} & \multicolumn{1}{p{12em}<{\centering}}{GCN-based} \\
%     \midrule
%     \textbf{Traffic Flow Prediction} & AutoEncoder~\cite{RNN} & ST-ResNet~\cite{STResNet}, ACFM~\cite{ACFM} & RNN~\cite{RNN}, Seq2Seq~\cite{DCRNN} & ASTGCN~\cite{ASTGCN}, MSTGCN~\cite{ASTGCN}, AGCRN~\cite{AGCRN} \\
%     \midrule
%     \textbf{Traffic Speed Prediction} & AutoEncoder~\cite{RNN} &       & RNN~\cite{RNN}, Seq2Seq~\cite{DCRNN} & DCRNN~\cite{DCRNN}, STGCN~\cite{STGCN}, GWNET~\cite{GWNET}, MTGNN~\cite{MTGNN}, TGCN~\cite{TGCN}, TGCLSTM~\cite{TGCLSTM} \\
%     \midrule
%     \textbf{On-Demand Service Prediction} & AutoEncoder~\cite{RNN} &       & RNN~\cite{RNN}, Seq2Seq~\cite{DCRNN} &  \\
%     \midrule
%     \textbf{Trajectory Next-Location Prediction} & FPMC~\cite{FPMC}  &       & ST-RNN~\cite{STRNN}, HST-LSTM~\cite{HSTLSTM}, ATST-LSTM~\cite{ATSTLSTM}, SERM~\cite{SERM}, DeepMove~\cite{DeepMove}, CARA~\cite{CARA}, LSTPM~\cite{Sun_Qian_Chen_Liang_Nguyen_Yin_2020} &  \\
%     \bottomrule
%     \end{tabular}%
%   \label{tab:implemented_model}%
% \end{table*}%

\subsection{Evaluation Module}
With the standardized data processing flow and prediction model interfaces, \name also offers standard evaluation procedures for spatial-temporal prediction tasks. Since the model output formats and evaluation metrics may vary across different spatial-temporal prediction tasks, \name develops specific evaluators for each task and supports various popular evaluation metrics.

\subsubsection{Evaluation Metrics} \
% \todo{For traffic state prediction task, \name supports commonly used value-based metrics, which includes Mean Absolute Error(MAE), Mean Squared Error(MSE), Rooted Mean Squared Error(RMSE), Mean Absolute Percent Error(MAPE), Coefficient of Determination($R^2$), and Explained variance Score(EVAR). For trajectory location prediction task, \name supports commonly used rank-based metrics, which includes Precision@K, Recall@K, F1-score@K, MRR@K (Mean Reciprocal Rank@K), and NDCG@K (Normalized Discounted Cumulative Gain@K)}

% For evaluation, the output of the first three tasks is real value, considered as \emph{regression tasks}, while the output of the fourth task is discrete value, considered as \emph{classification tasks}. Hence, we conduct different evaluation settings for the two kinds of tasks. Next, we describe two important aspects related to the evaluation module.  

\textbf{Metrics for Regression Tasks:} 
In \name, regression tasks consist of Traffic Flow Prediction, Traffic Speed Prediction, On-Demand Service Prediction, Origin-Destination Matrix Prediction, Traffic Accidents Prediction, and Travel Time Prediction. These tasks output real numbers and are evaluated using commonly used value-based metrics, which include Mean Absolute Error (MAE), Mean Squared Error (MSE), Root Mean Squared Error (RMSE), Mean Absolute Percentage Error (MAPE), Coefficient of Determination ($R^2$), and Explained Variance Score (EVAR). Their calculation formulas for these metrics are as follows:
\begin{equation}\small
    MAE=\frac{1}{n}\sum_{i=1}^n|\hat{y_{i}}-y_i|
\end{equation}
\begin{equation}\small
    MSE=\frac{1}{n}\sum_{i=1}^n(\hat{y_{i}}-y_i)^2
\end{equation}
\begin{equation}\small
    RMSE=\sqrt{\frac{1}{n}\sum_{i=1}^n(\hat{y_{i}}-y_i)^2}
\end{equation}
\begin{equation}\small
    MAPE=\frac{1}{n}\sum_{i=1}^n|\frac{\hat{y_{i}}-y_i}{y_i}|*100\%
\end{equation}
\begin{equation}\small
    R^2=1-\frac{\sum_{i=1}^n(y_i-\hat{y_i})^2}{\sum_{i=1}^n(y_i-\bar{y})^2}
\end{equation}
\begin{equation}\small
    EVAR =1-\frac{Var(y_i-\hat{y_i})}{Var(y_i)}
\end{equation}
where $\boldsymbol{y}=\{y_1,y_2,...,y_n\}$ is the ground-truth value, $\hat{\boldsymbol{y}} = \{\hat{y_1}, \hat{y_2}, ..., \hat{y_n}\}$ is the prediction value, $n$ is the number of samples, $\bar{y}=\frac{1}{n}\sum_{i=1}^ny_i$ is the mean value, $Var(y_i)=\frac{1}{n}\sum_{i=1}^n(y_{i}-\bar{y})^2$ is the variance.

\textbf{Metrics for Classification Tasks:} 
The classification task in \name is Trajectory Next-Location Prediction. The output of this task in \name is a probability distribution over the candidate's next locations. The Trajectory Next-Location Prediction task is evaluated using various rank-based metrics, including Precision@K, Recall@K, F1-score@K, MRR@K (Mean Reciprocal Rank@K), and NDCG@K (Normalized Discounted Cumulative Gain@K). The calculation formulas for these metrics are as follows:

\begin{equation}\small
    Precision@K=\frac{\sum_{i=1}^{N}|\operatorname{Hit}(i)|}{N \times K}
\end{equation}
\begin{equation}\small
    Recall@K=\frac{\sum_{i=1}^{N}|\operatorname{Hit}(i)|}{N}
\end{equation}
\begin{equation}\small
    F1@K=\frac{2 \times \text { Precision@ } \times \text { Recall@ } K}{\text { Precision } @+\text { Recall@ } K}
\end{equation}
\begin{equation}\small
    MRR@K=\frac{1}{N} \sum_{i=1}^{N} \frac{1}{\operatorname{Rank}(i)}
\end{equation}
\begin{equation}\small
    NDCG@K=\frac{1}{N} \sum_{i=1}^{N} \frac{1}{\log _{2}(\operatorname{rank}(i)+1)}
\end{equation}
where $N$ is the number of test data, $i$ is the $i$-th test data, $K$ is the top $K$ prediction outputs for evaluation, $T(i)$ is the real next hop position in the $i$-th test data, $R(i)$ is the set of the top K locations in the prediction result of the $i$-th test data, $Hit(i)$ is the set of predicted hit locations in the $i$-th test data, which means $T(i) \cap R(i)$, $Rank(i)$ is the ranking of $T(i)$ in $R(i)$ in the $i$-th test data, and $|*|$ is the modulo operator of a set.

\textbf{Metrics for Fundamental Tasks:}
Fundamental tasks in \name provide support for macro and micro prediction tasks, including map matching and road network representation learning. The road network representation task requires combining with specific downstream tasks to evaluate the performance of the representation vectors. For instance, if a road segment classification task is used for evaluation, it is a classification task. If a road flow prediction task is used for evaluation, it is a regression task. The evaluation metrics for these tasks are similar to the ones described above. Here we focus on the evaluation metrics for the map matching task.

\name evaluates the map matching task using three metrics: RMF (Route Mismatch Fraction), AN (Accuracy in Number), and AL (Accuracy in Length), which have been used in previous works such as \cite{HMMM, IVMM}.
\begin{equation}\small
    RMF= \frac{d_{-}+d_+}{d_0}
\end{equation}
\begin{equation}\small
    AN=\frac{\#Rc}{\#R}
\end{equation}
\begin{equation}\small
    AL=\frac{\sum len(Rc)}{\sum len(R)}
\end{equation}
where $d_0$ denotes the length subtracted from the error, $d_+$ denotes the length added to the error, $d_-$ is the total length of the real path, $Rc$ denotes correctly matched roads, $R$ denotes all roads of the real route. $len()$ denotes the length of a set of roads.

\subsubsection{Evaluation Strategies} To evaluate the performance of spatial-temporal prediction models in a flexible manner, \name offers two main strategies.

Firstly, users can divide the dataset into training, validation, and testing sets with a ratio of their choice. The training set is used for model training, while the validation set is used for hyper-parameter tuning and preventing overfitting. The testing set is used to evaluate the final performance of the trained model.

Secondly, \name allows users to set different window sizes for evaluation. For macro group prediction tasks, users can set various input and output time windows, and \name will partition the input data based on the window size, enabling multi-step predictions using historical observations of different lengths. For micro individual prediction tasks, trajectories are split based on window settings, with options for time-based or length-based windows. Users can set the window size and type to evaluate the model's performance on trajectories of varying lengths, such as long, medium, and short trajectories.

These two strategies can be combined with various evaluation metrics to perform comprehensive evaluations of models for the same task, providing greater flexibility and adaptability in assessing model performance.

\subsection{Execution Module}
The execution module in \name serves as the central hub that controls the interactions between other modules to facilitate model training and performance evaluation. Users can modify the parameter settings of this module to adjust the effect of model training. It supports various model training strategies that optimize the model and includes a built-in automatic hyper-parameter tuning module to reduce the user's workload and achieve automatic optimization of the model.

\subsubsection{Model Training Techniques} \name offers various training techniques to train deep neural networks effectively. These techniques can be customized by modifying the parameter settings of the execution module. Here are some of the techniques supported by \name:
\begin{itemize}
    \item \textit{Optimizer options}: During the training of a deep learning model, the optimizer is mainly used to update the parameters of the network in order to minimize the loss function. \name supports several optimizers including SGD~\cite{SGD}, RMSProp~\cite{RMSProp}, Adam~\cite{Adam}, and AdaGrad~\cite{AdaGrad}. Different optimization algorithms have different ways of updating the network parameters and are better suited to different scenarios. Users can choose the appropriate optimization algorithm for their usage scenarios to achieve better training results.
    \item \textit{Learning rate adjustment strategies}: Learning rate is a key parameter of neural network models, which controls the speed of gradient-based adjustment of network weights and determines the convergence of the loss function to the optimal solution. % \name supports six learning rate adjustment strategies including StepLR~\cite{pytorch_steplr}, MultiStepLR~\cite{pytorch_multisteplr}, ExponentialLR~\cite{pytorch_exponentiallr}, LambdaLR~\cite{pytorch_lambdalr}, and ReduceLROnPlateau~\cite{pytorch_reducelronplateau}. These strategies help the model converge to the optimal solution more easily in multiple iterations.
    \item \textit{Loss Function options}: Various loss functions are used in deep learning and different loss functions compute the loss in different ways to obtain different training results. \name supports five types of loss functions including Cross-entropy Loss, L1 Loss, L2 Loss, Huber Loss~\cite{huber1992robust}, LogCosh Loss~\cite{logcosh}, and Quantile Loss~\cite{koenker2001quantile}.
    \item \textit{Early Stopping}: Overfitting may occur as the number of training rounds increases. \name supports the early stopping method to prevent overfitting. Users can specify whether to use the early stop mechanism and the size of the duration rounds through parameter configuration.
    \item \textit{Gradient Clipping}: Gradient explosion may occur during training. To avoid this, \name supports gradient clipping strategies~\cite{gradientcropping}. Users can specify whether to use gradient clipping through parameter configuration.
\end{itemize}

\subsubsection{Automatic Hyper-parameter Tuning}
Hyper-parameter tuning has a significant impact on the performance of deep learning models. To ease the burden of manual parameter tuning, \name provides an automatic hyper-parameter tuning mechanism. We implement this feature using the third-party library \emph{Ray Tune}\cite{liaw2018tune}, which supports various search algorithms such as Grid Search, Random Search, and Bayesian Optimization. Users can specify the parameters to be tuned and their search space in a configuration file and select the tuning method. \name will then sample multiple times from the search space and run the model in a distributed manner, automatically saving the best parameter values and corresponding model prediction results. An example of automatic hyper-parameter tuning is presented in Section~\ref{tuning}.

\subsection{Configuration Module}
\name utilizes the configuration module to set the parameters for the entire framework. The experiment parameter configuration is determined by three factors: parameters passed from the command line, user-defined configuration file, and default configuration files of the modules in \name. The priority of the above three parameter configuration methods decreases in order, with higher priority parameters overriding lower priority parameters with the same name. With this priority design, users can flexibly adjust the parameter configuration of an experiment through the first two methods.

To adjust parameters through the command line, users only need to use "\lstinline{--parameter_name}" when running \name. However, it is important to note that only frequently adjusted parameters in an experiment, such as batch size and learning rate, are allowed to be passed from the command line. In order to allow users to modify default parameters more extensively, \name allows users to pass the name of a user-defined configuration file through the command line, which is then read by the system to set the parameter configuration.

\subsection{Experiment Management and Visualization Platform} 

% \begin{figure}[t]
%     \vspace{-1mm}
%     \centering
%     \includegraphics[width=0.8\columnwidth]{figures/metrla.jpg}
%     \caption{\todo{Visualization of METR\_LA dataset}}
%     \label{fig:metr}
%     \vspace{-0.4cm}
% \end{figure}

\begin{figure}[t]
  \centering
  \subfigure[Platform Homepage]{
    \includegraphics[width=0.45\columnwidth]{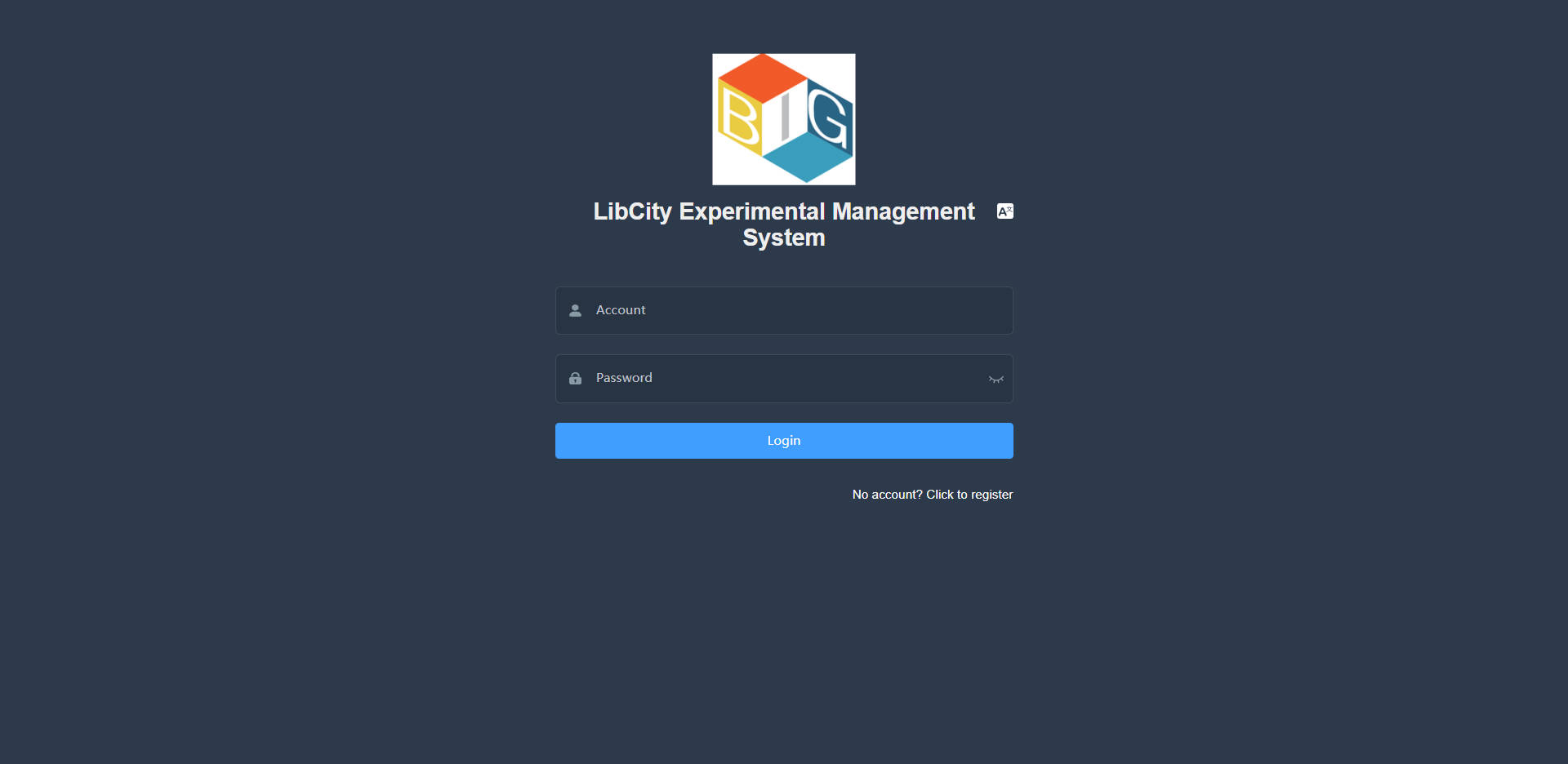}
    \label{fig:p1}
  }
  \subfigure[Visualization of Beijing Flow Dataset]{
    \includegraphics[width=0.45\columnwidth]{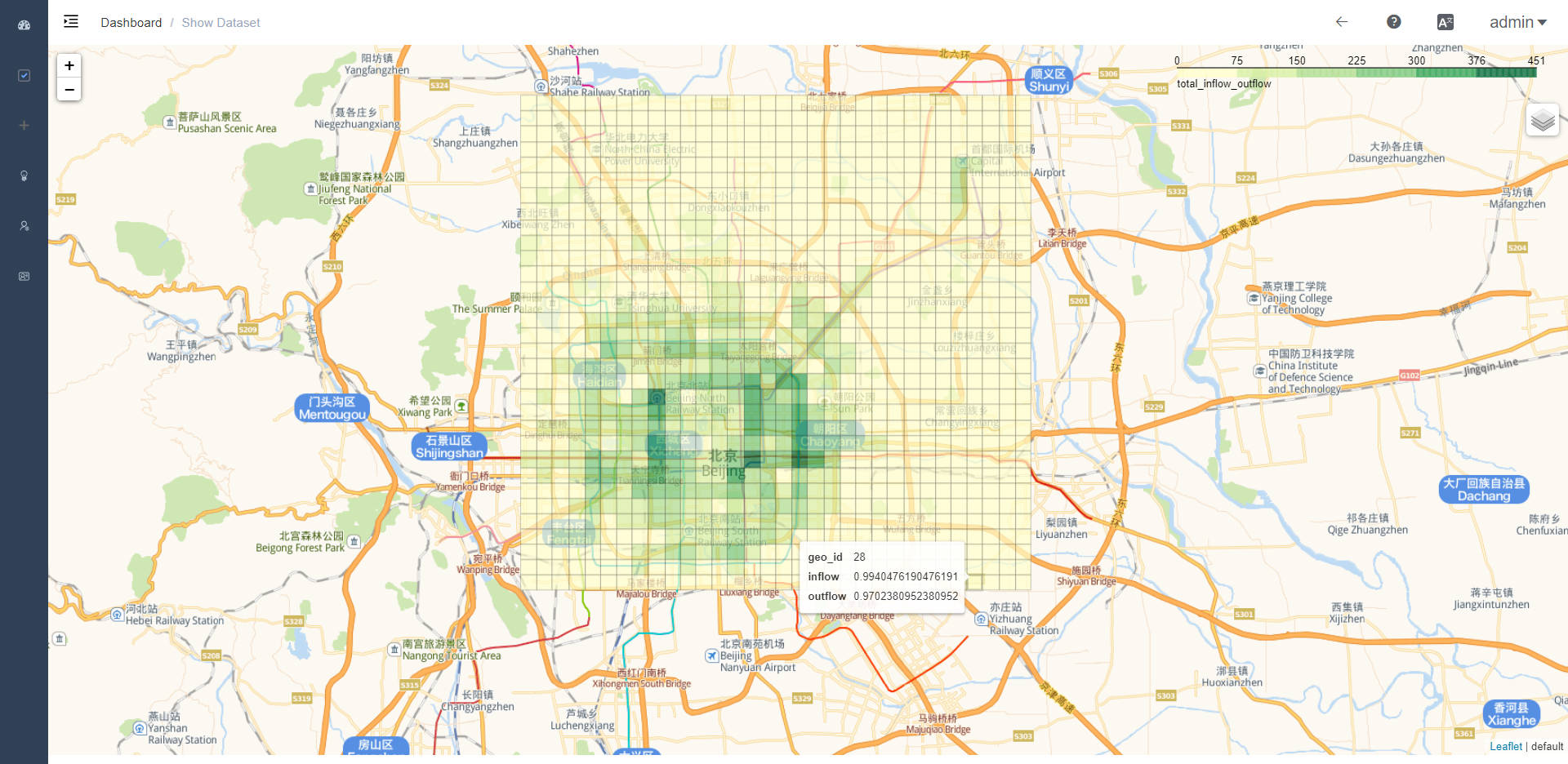}
    \label{fig:p2}
  }
  \subfigure[Create Experiment]{
    \includegraphics[width=0.45\columnwidth]{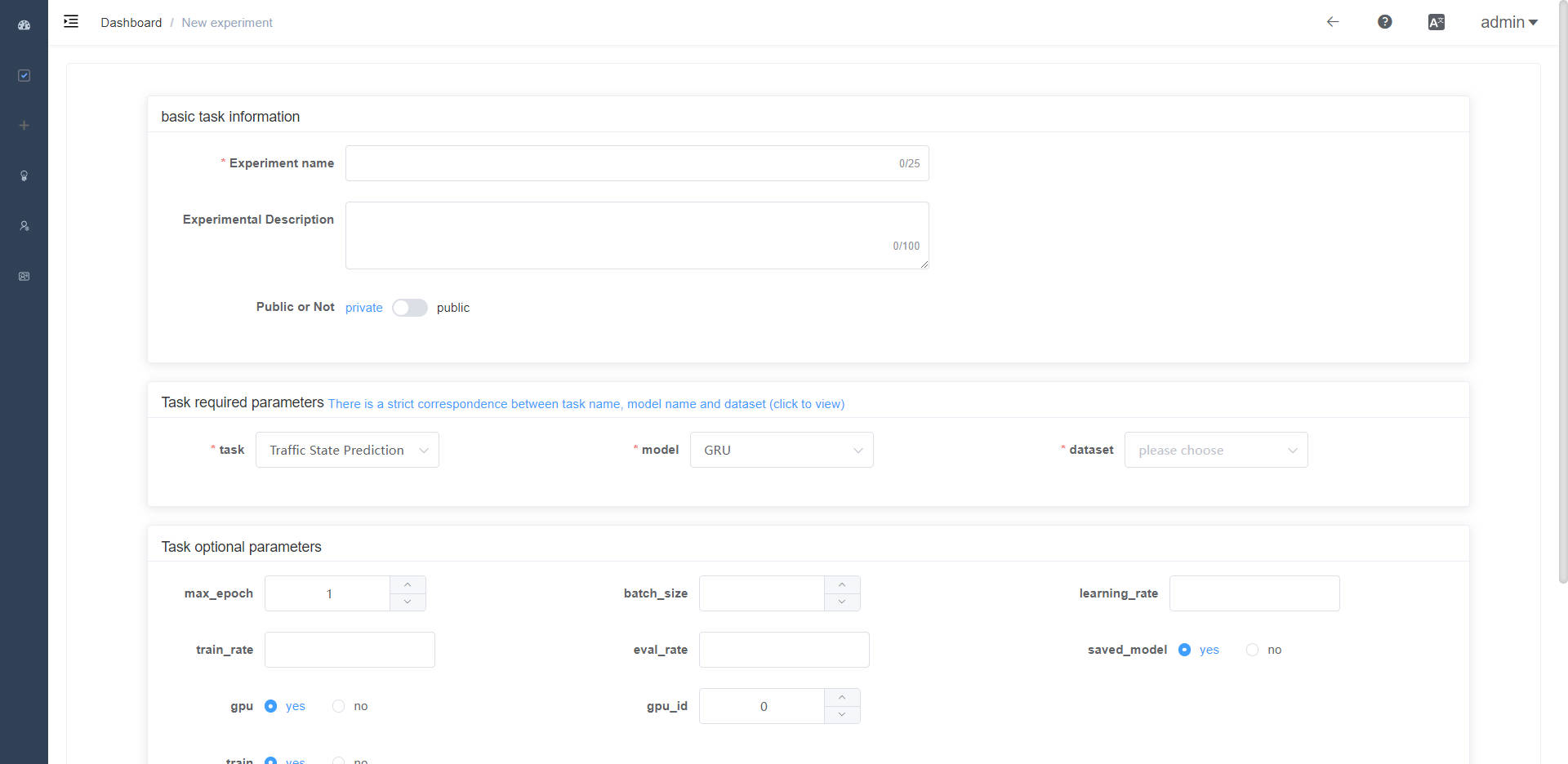}
    \label{fig:p3}
  }
  \subfigure[Model Performance Comparison]{
    \includegraphics[width=0.45\columnwidth]{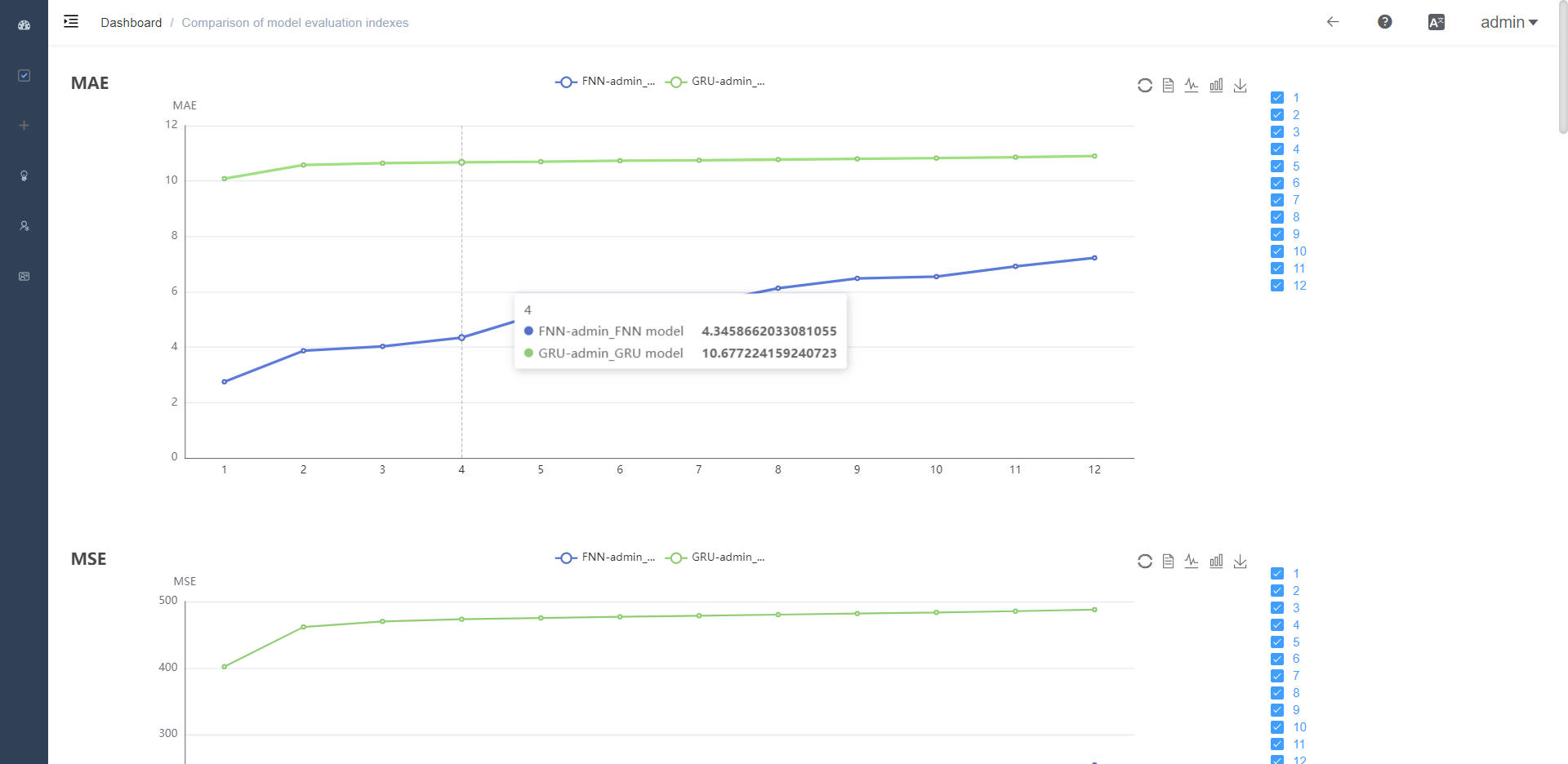}
    \label{fig:p4}
  }
  \caption{Experiment Management and Visualization Platform}
  \label{fig:platform}
  \vspace{-0.3cm}
\end{figure}

We have created a web-based experiment management and visualization platform for convenient experimentation with \name's models and datasets as shown in Figure~\ref{fig:p1}. The platform features a user-friendly graphical interface and comprehensive functionalities to support spatial-temporal prediction research. Users can easily upload new datasets and visualize them, as demonstrated in Figure~\ref{fig:p2}, which shows the visualization of the Beijing traffic flow dataset. After configuring the model's parameters, users can create new experiments through a straightforward web interface, as shown in Figure~\ref{fig:p3}. They can then execute the experiments at their convenience, with the training logs available for viewing during the experiment. After the experiment's execution, users can obtain evaluation results of the model on specific metrics and visualize the prediction results. Additionally, the platform offers an experimental comparison function, allowing users to compare the performance of different models shown in Figure~\ref{fig:p4}.

We have built the experiment management and visualization platform using Django\footnotemark[10], Vue\footnotemark[11], and MySQL\footnotemark[12]. The platform's open-sourced code can be found on GitHub\footnotemark[13].

\footnotetext[10]{\url{https://www.djangoproject.com}}
\footnotetext[11]{\url{https://vuejs.org}}
\footnotetext[12]{\url{https://www.mysql.com}}
\footnotetext[13]{\url{https://github.com/LibCity/Bigscity-LibCity-WebTool}}

\section{Usage Examples of \name} \label{usage}

This section provides several examples of using \name to assist users in getting started with the framework. The examples cover running existing models, conducting automatic parameter tuning, and adding new models to \name.

% \begin{figure}[t]
%     \vspace{-1mm}
%     \centering
%     \includegraphics[width=0.8\columnwidth]{figures/run.png}
%     \caption{\todo{Existing Model Run Flow}}
%     \label{fig:run}
%     \vspace{-0.4cm}
% \end{figure}

% It involves downloading and formatting the raw dataset into atomic files, configuring the framework either through the command line, a configuration file, or using the default parameters, pre-processing and splitting the dataset according to the user's settings, initializing the model based on the task and model name selected by the user, and finally, training and evaluating the model on the specified dataset.
\subsection{Running Existing Models } 
The general process of running existing models in \name is listed as follows: 

\begin{enumerate}[label=\roman*)]
    \item Dataset formatting: Users must download and convert the raw dataset into atomic files.
    \item Configuration setup: Users can configure the experiment parameters through a configuration file, the command line, or the default parameters of \name. The configuration module's parameters form the foundation of the entire framework.
    \item Dataset pre-processing and splitting: Based on the user's parameter settings, \name pre-processes and splits the dataset. Users can specify custom dataset division ratios and data pre-processing thresholds. For example, users may filter out trajectories with lengths less than 5.
    \item Model initialization: \name creates a model object based on the user's task and model name selection.
    \item Training and evaluation: Once the data and model are ready, \name calls the executor module to train and evaluate the model on the specified dataset. The training logs, trained models, and model evaluation results are automatically saved for the user's use.
\end{enumerate}

The above steps are performed automatically by \name's unified entry file (\textit{run\_model.py}). Users need only execute a single command in the command line to initiate the model running process. Three essential parameters must be specified at runtime, namely the task, model, and dataset, via \textit{-{}-task}, \textit{-{}-model}, and \textit{-{}-dataset} options, respectively. For example, running the GRU model on the METR\_LA dataset for 50 epochs on the second GPU block requires the following command:

\textit{python run\_model.py -{}-task traffic\_state\_prediction -{}-model GRU -{}-dataset METR\_LA -{}-gpu 2 -{}-epoch 50}
% \begin{verbatim}
% python run_model.py --task traffic_state_prediction --model GRU --dataset METR_LA --gpu 1 --epoch 50
% \end{verbatim}

% \begin{tcolorbox}[colback=gray!10,colframe=gray!70]
% \lstinline{python run_model.py --task traffic_state_pred --model GRU --dataset METR_LA --gpu_id 2 --max_epoch 50}
% \end{tcolorbox}

% During the model running process, the logs of the training progress, the trained models, and the evaluation results are automatically saved for the user's use.

% This simple and flexible command line interface reflects the ease of use and flexibility of \name. Moreover, users can modify the parameter settings at runtime through the command line or configuration file, demonstrating the flexibility and extensibility of \name.

\subsection{Running Automatic Parameter Tuning}  \label{tuning}

Considering that hyper-parameter tuning significantly impacts the performance of deep learning models, \name has introduced an automatic hyper-parameter tuning mechanism that can easily optimize a given model based on the user-provided hyper-parameter range. The general steps for running the automatic tuning function are as follows:

\begin{enumerate}[label=\roman*)]
    \item Setting the hyper-parameter search space. Users must specify the hyper-parameters to be tuned and their corresponding value ranges in a JSON file. For example, the search space for the hidden layer dimension parameter can be a discrete set of categorical variables, such as [50, 100, 200]. The user can specify the file name of the parameter space file using the parameter \textit{-{}-space\_file}.
    \item Selecting the tuning method. \name uses the third-party library \emph{Ray Tune}~\cite{liaw2018tune} to implement automatic hyper-parameter tuning, which supports various search algorithms, including Grid Search, Random Search, and Bayesian Optimization. The user can select the tuning method by specifying the parameter \textit{-{}-search\_alg}.
    \item Starting the tuning process. Users can execute a single command in the command line to automatically tune the model parameters. During the tuning process, \name will sample the corresponding parameter values from the search space and perform model training and validation. After all the samples are validated, the script will output the best parameter combination on the terminal and save them to a log file. Here is an example of the command to run the automatic tuning function: \textit{python hyper\_tune.py -{}-task traffic\_state\_pred -{}-model GRU -{}-dataset METR\_LA -{}-search\_alg BasicSearch -{}-space\_file sample\_space\_file}

% \begin{tcolorbox}[colback=gray!10,colframe=gray!70]
% \lstinline{python hyper_tune.py --task traffic_state_pred --model GRU --dataset METR_LA}
% \lstinline{--search_alg BasicSearch --space_file sample_space_file}
% \end{tcolorbox}

\end{enumerate}

\subsection{Implementing a New Model} 
The modular design and unified interface definition of \name provide flexibility for user-defined extensions and excellent scalability. Developing a new spatial-temporal prediction model using \name is straightforward. The general process for developing a new model using \name is as follows:

\begin{enumerate}[label=\roman*)]
    \item Create a new model file and define a model class that inherits from the existing model abstraction class provided by \name. All \taskcnt tasks supported by \name have their abstract classes implemented.
    \item Implement the \textit{\_\_init\_\_()} function to initialize the model according to the configuration parameters. The input parameter \textit{config} contains the configuration information.
    \item Implement the \textit{calculate\_loss()} function to calculate the loss between the predicted result and the true value during model training. The goal of model training is to optimize this loss.
    \item Implement the \textit{predict()} function to return the model prediction results during prediction.
    \item Configure the default parameter configuration file for the new model to specify the required parameters and their values for running the model.
    \item Run the model and evaluate its performance on the selected dataset.
\end{enumerate}

Thus, developing new models using \name is simplified by focusing on implementing only three interfaces, while \name handles other details like data splitting, model training, and performance evaluation. This design approach simplifies the development process of new models and highlights the scalability of \name.

% \subsection{Visualize Atomic Files} 
% \name offers many useful features for researchers to enhance its overall functionality and usability. One of them is visualization of atomic files. The process of visualizing the atomic files in \name is listed as follows:
% \begin{enumerate}[label=\roman*)]
%     \item Download and convert the raw dataset into atomic files.
%     \item Run \textit{visualize.py} script to convert atomic files into the \textit{GeoJSON}~\footnote{\url{https://tools.ietf.org/html/rfc7946\#section-1}} format, \ie \textit{Point}, \textit{LineString}, \textit{Polygon} etc.
%     \item Visualization of Geojson files using GIS software, e.g. QGIS~\footnote{\url{https://www.qgis.org/id/site/}}, ArcGIS~\footnote{\url{https://www.arcgis.com/index.html}}, etc.
% \end{enumerate}

% For example, the following command can be used to perform the atomic file visualization of METR\_LA dataset:

% \textit{python visualize.py --dataset METR\_LA}

% \begin{figure}[t]
%     \vspace{-1mm}
%     \centering
%     \includegraphics[width=1\columnwidth]{figures/metrla.jpg}
%     \caption{Visualization of METR\_LA dataset}
%     \label{fig:metr}
%     \vspace{-0.4cm}
% \end{figure}

% Figure~\ref{fig:metr} shows the heat map obtained using ArcGIS to visualize the METR\_LA dataset, with darker colors indicating faster speed at that location.

% \input{experiments.tex}

\section{Comparison with Existing Libraries} \label{cmp}

\begin{table*}[t]
\small
    \centering
    \caption{Comparison with existing libraries by 2023/4/22}
        \begin{tabular}{c|c|c|c|c|c|c}
        \toprule
           {Framework\textbackslash{}Metric} & {Modularization}  & {\#Fork }& {\#Star }& {\#Models} & {\#Datasets} & {\#Issues }\\ 
        \midrule
        DL-Traff & Low  & 34 & 179 & 18 & 7 & 2/3 \\ 
        DGCRN & Low & 64 & 173 &  12 & 3 & 9/13  \\ 
        FOST & Middle  & 43 & 208 & 6 & 0 & 4/13 \\ 
        % Pytorch Geometric Temporal & High  & \textbf{287} & \textbf{2002} & 20 & 12 & 23/145 \\ 
        \textbf{\name} & \textbf{High} & \textbf{111} & \textbf{485} & \textbf{\modelCnt} & \textbf{\rawDataCnt} & \textbf{0/73} \\ 
        \bottomrule
        \end{tabular}
    \label{tab:framework_cmp}
\end{table*}

To the best of our knowledge, \name~\cite{libcity} is the first spatial-temporal prediction library that enables researchers to conduct comprehensive comparative experiments and develop new models. Recently, other researchers have proposed benchmarks in spatial-temporal prediction similar to \name. This section compares with others to demonstrate the advantages of \name.

DL-Traff~\cite{jiang2021dl} is an open-source project offering a traffic prediction benchmark using grid-based and graph-based models. DGCRN~\cite{DGCRN} summarizes previous work and produces a benchmark in traffic prediction. However, these two projects only accumulate the model codes from past research work without a modular design, which makes it inconvenient for users to use. Furthermore, they only focus on macro-level traffic prediction without contributing to micro-level individual prediction tasks. Microsoft FOST~\footnotemark[14] (Forecasting Open Source Tool) is a general forecasting tool that aims to provide an easy-to-use tool for spatial-temporal forecasting, but its supported models and applications are limited.

\footnotetext[14]{\url{https://github.com/microsoft/FOST}}

Table~\ref{tab:framework_cmp} illustrates that \name outperforms the other compared tools in terms of the number of models and datasets it can handle. Furthermore, \name designs a unified storage format for spatial-temporal data, which is a great help to promote the standardization of the field. The modular design of \name allows for scalability and enables developers to easily create new models using its pipeline. The high number of Stars and Forks indicates that \name is popular among the open-source community. The number of open issues and total issues also suggests that the developers of \name are active in the community, which helps to address user inquiries and promote further development in the field.

In addition, it is worth noting that numerous similar experimental libraries are available in other research fields. For instance, RecBole~\cite{recbole} is a recommendation algorithm framework that reproduces a vast range of recommendation models and provides various evaluation strategies and data preprocessing operations, making it easy to conduct experiments. Meanwhile, MMDetection~\cite{mmdetection} adopts a modular design, enabling researchers to efficiently develop new models based on it for object detection tasks. FastReID~\cite{he2020fastreid} continuously reproduces state-of-the-art models and releases corresponding pre-trained models for both research and industrial purposes.

\name combines the strengths of the libraries as mentioned above, such as various baseline models, diverse evaluation strategies, and modular design. As a result, it not only facilitates researchers to conduct experiments and develop new models but also promotes standardization within the spatial-temporal prediction field.

% PyTorch Geometric Temporal~\cite{rozemberczki2021pytorch} is a temporal (dynamic) extension library for PyTorch Geometric. This work only focuses on the application of spatial-temporal prediction models related to graph neural networks, ignoring others.
 
% Pytorch Geometric Temporal is more well-liked than \name, which may be caused by the popularity of Pytorch. However, \name classifies different spatiotemporal prediction tasks more thoroughly. We think that as the framework continuously improves, more R\&D personnel will be drawn to engage in it.

\section{Conclusion} \label{conclusion}

In this work, we present a comprehensive review of urban spatial-temporal prediction and proposes a unified storage format for spatial-temporal data, called atomic files. Building on this, we introduce \name, a unified and comprehensive open-source library for urban spatial-temporal prediction that includes \rawDataCnt spatial-temporal datasets and \modelCnt spatial-temporal prediction models covering \taskcnt mainstream sub-tasks of urban spatial-temporal prediction. By conducting extensive experiments using \name, we establish a comprehensive model performance leaderboard that identifies promising research directions for spatial-temporal prediction.

To the best of our knowledge, \name is the first open-source library for urban spatial-temporal prediction, providing a valuable tool for exploring and developing spatial-temporal prediction models. We will continuously expand \name to contribute to the spatial-temporal prediction field in the future. For example, we can cover more spatial-temporal prediction tasks, such as climate prediction, air quality prediction, theft prediction, etc.

% is not selected <OR> if conference mode is selected - because compsoc
% conference papers position the abstract like regular (non-compsoc)
\IEEEdisplaynontitleabstractindextext

% \fi
\IEEEpeerreviewmaketitle

\ifCLASSOPTIONcaptionsoff
  \newpage
\fi



%\bibliographystyle{IEEEtran}

% H.~Kopka and P.~W. Daly, \emph{A Guide to {\LaTeX}}, 3rd~ed.\hskip 1em plus

% \end{thebibliography}
\bibliographystyle{IEEEtran}
\bibliography{references}


% \end{IEEEbiography}


% % insert where needed to balance the two columns on the last page with
% %\newpage



%\vfill


\end{document}